\pdfoutput=1
\PassOptionsToPackage{dvipsnames}{xcolor}
\documentclass{article} 
\usepackage{iclr2027_conference,times}

\usepackage{amsmath,amsfonts,bm}

\def\eqref#1{equation~\ref{#1}}
\def\Eqref#1{Equation~\ref{#1}}

\def\1{\bm{1}}

\DeclareMathAlphabet{\mathsfit}{\encodingdefault}{\sfdefault}{m}{sl}
\SetMathAlphabet{\mathsfit}{bold}{\encodingdefault}{\sfdefault}{bx}{n}

\usepackage[dvipsnames]{xcolor}
\usepackage{hyperref}
\usepackage{etoc} 

\usepackage{url}
\usepackage{fontawesome5}
\usepackage{booktabs}
\usepackage{enumitem}      

\usepackage{amssymb}          
\usepackage{pifont}           
\usepackage[outline]{contour}
\usepackage{wrapfig}
\usepackage{algorithm}
\usepackage{algpseudocode}
\usepackage{multirow}

\usepackage{tabularx}

\contourlength{0.045em}
\definecolor{goldedge}{HTML}{C98500}

\DeclareRobustCommand{\retrainstar}{\contour{black}{\textcolor[HTML]{EDA100}{\ding{72}}}}   
\DeclareRobustCommand{\targetstar}{{\contourlength{0.09em}\contournumber{16}%
  \contour{goldedge}{\textcolor{white}{\ding{72}}}}} 
\DeclareRobustCommand{\origdiamond}{\contour{white}{\ding{117}}}                            
\DeclareRobustCommand{\pintdot}{\contour{black}{\textcolor[HTML]{2A78D6}{\ding{108}}}}      

\hypersetup{
    colorlinks=true,
    citecolor=Blue,
    linkcolor=RoyalBlue,
    urlcolor=RoyalBlue
}
\usepackage{subcaption}

\usepackage{amsthm}
\usepackage{mathtools}
\newtheorem{theorem}{Theorem}[section]

\newtheorem{proposition}[theorem]{Proposition}

\theoremstyle{definition}
\newtheorem{definition}[theorem]{Definition}

\usepackage{graphicx}

\title{\centering Forget What Comes Next:\\ LLM Unlearning via Statistical Independence}

\author{%
\hspace*{-0.2in}\parbox[t]{\textwidth}{\centering
\textbf{Andrii Shkabrii}$^{1,2}$\hspace{3em}\textbf{Claudia Plant}$^{1,3}$\hspace{3em}\textbf{Martin Pawelczyk}$^{1}$\\[6pt]
\normalfont $^{1}$Faculty of Computer Science, University of Vienna \\ 
\normalfont $^{2}$Doctoral School Computer Science, University of Vienna \\ 
\normalfont $^{3}$Research Network Data Science, University of Vienna} 
}

\iclrfinalcopy 
\begin{document}
\etocdepthtag.toc{mtmain}   

\maketitle
\lhead{Preprint.}

\begin{abstract}
Large language models (LLMs) can memorize and reproduce fragments of their training data, raising privacy and legal concerns that motivate machine unlearning. Yet poisoned training examples can remain statistically detectable after unlearning, which raises the question of whether current unlearning objectives remove what the model has memorized at all. In this work, we introduce \textsc{PINT}, a novel algorithm for machine unlearning that uses the Hilbert-Schmidt Independence Criterion (HSIC) to reduce the statistical dependence between tokens to be forgotten and the internal representations used to predict them. We formally relate our HSIC-based objective to Exact Memorization through a computable upper bound, connecting the training objective directly to the model's ability to reproduce the forgotten tokens. In extensive experiments, we compare forgetting-utility trade-offs across eight settings with models up to 7B parameters and demonstrate scalability to 70B parameter models. Our evaluations cover unlearning Gaussian poisons, personal information, hazardous knowledge in biology and cybersecurity, and copyrighted books and news articles. \textsc{PINT} achieves the strongest forgetting-utility trade-off on average and is the only method to make Gaussian poisons undetectable while maintaining near-original model utility.
\end{abstract}

\begin{center}
  \small
  \href{https://github.com/shkabrii/PINT}
       {\textcolor{black}{\faGithub}\ \texttt{Codebase}}
\end{center}

\section{Introduction}
\begin{wrapfigure}{r}{0.44\linewidth}
\centering
\vspace{-1\baselineskip}
\includegraphics[width=\linewidth]{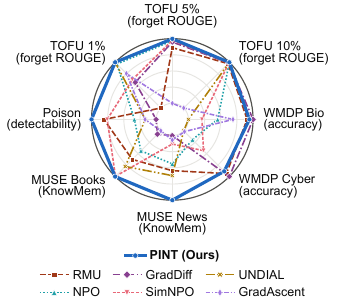}
\caption{\textbf{\textsc{PINT} achieves the strongest trade-off between forgetting and utility.} Each axis is one benchmark, on which methods are ranked by the area under their forgetting-utility frontier, from the worst on the inner circle to the best on the outer one.}
\vspace{-12pt}
\label{fig:pareto-gw}
\end{wrapfigure}

The tendency of large language models (LLMs) to memorize and reproduce fragments of their training data, including personally identifiable information and copyrighted text~\citep{carlini2021extracting, carlini2022quantifying}, poses significant privacy and legal concerns. With the \textit{right to be forgotten} now formalized in major data protection frameworks across the EU, the US, and China~\citep{gdpr2016, ccpa2018, pipl2021}, there is a growing need for methods that can remove the influence of specific training examples without the prohibitive cost of retraining from scratch. A growing body of work on approximate \textit{machine unlearning} has made empirical progress in this direction~\citep{jang2023knowledge, tofu, wmdp-rmu, npo}. Nevertheless, significant gaps remain: \citet{watermarks} show that current methods may fail to unlearn poisoned examples injected during training, leaving them statistically detectable even after an unlearning attempt. 

To understand how an LLM can forget a memorized sequence, we look at how it predicts each token from the preceding context. At each position in a sequence, an autoregressive LLM encodes the preceding context into a prefix representation. From this representation, the output head predicts the next token. Reproducing a memorized sequence requires that these prefix representations carry information about the tokens that follow. This relationship can be expressed in terms of statistical dependence. So, can we induce forgetting by minimizing this statistical dependence? 

\begin{figure}[t]
\centering
\includegraphics[width=\textwidth]{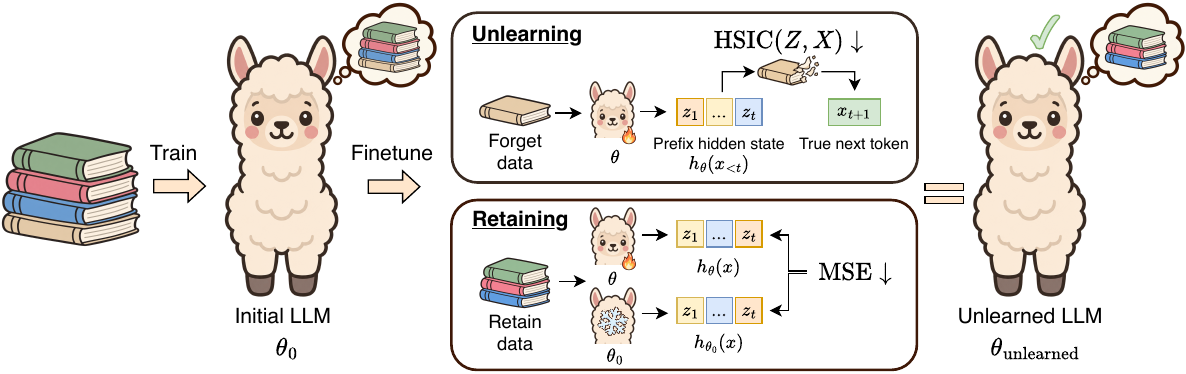}
\caption{\textbf{Overview of \textsc{PINT}.} On the \textit{forget set}, \textsc{PINT} minimizes the HSIC between prefix representations and next tokens, reducing their statistical dependence. On the \textit{retain set}, it constrains prefix representations to stay close to those of the original model, preserving general model utility.}
\label{fig:overview}
\end{figure}

To answer this, we propose \textsc{PINT} (\textbf{P}refix-representation \textbf{I}ndependence of the \textbf{N}ext \textbf{T}oken). 
On the data targeted for removal (the \textit{forget set}), \textsc{PINT} minimizes the Hilbert-Schmidt Independence Criterion (HSIC)~\citep{hsic} between prefix representations and their corresponding next tokens, capturing and penalizing complex, non-linear dependencies. 
To preserve utility, \textsc{PINT} constrains representations on a \textit{retain set} to remain close to those of the original model (see Figure~\ref{fig:overview} for an overview).
As illustrated in Figure~\ref{fig:pareto-gw}, this approach is highly effective: across benchmarks for personal information, hazardous knowledge, copyrighted text and Gaussian poisons, \textsc{PINT} achieves the strongest forgetting-utility trade-off on average. It is also the only method that makes Gaussian poisons statistically undetectable~\citep{watermarks} while retaining near-original utility.

In addition to our empirical evidence, we formally connect the unlearning objective to memorization. Exact Memorization (EM)~\citep{tirumala2022memorization} measures how many tokens of the forgotten text a model predicts correctly. We prove that EM admits an HSIC-based upper bound and measure this bound on LLMs with up to 7B parameters (Figure~\ref{fig:theory-verification}). To the best of our knowledge, this is the first work to use HSIC between prefix representations and next tokens as an LLM unlearning objective.

Our main contributions are as follows:
\begin{enumerate}[label=\(\bullet\), leftmargin=5mm,nosep] 
\itemsep0.15cm
    \item We introduce \textbf{\textsc{PINT}, an LLM unlearning method based on statistical independence}. On the \textit{forget set}, it makes prefix representations independent of their next tokens by minimizing HSIC, while keeping \textit{retain set} representations close to those of the original model.

    \item We derive an \textbf{HSIC-based upper bound on Exact Memorization}, formally connecting the \textsc{PINT} objective to how much of the forgotten text the model can still reproduce.

    \item \textsc{PINT} achieves \textbf{the strongest forgetting-utility trade-off} across eight unlearning settings with models up to 7B parameters, covering personal information, hazardous knowledge, copyrighted text and Gaussian poisons. It is the only method that makes Gaussian poisons statistically undetectable while retaining near-original model utility.
\end{enumerate}

\section{Related Work}
\textbf{LLM unlearning.} One approach to approximate LLM unlearning is to reduce the likelihood assigned to \textit{forget set} responses. Gradient ascent does this directly by minimizing their log-likelihood, while gradient difference~\citep{graddiff} additionally maximizes the likelihood of \textit{retain set} responses to preserve utility. Negative Preference Optimization (\textsc{NPO})~\citep{npo} reduces \textit{forget set} likelihood relative to the original model. Simple NPO (\textsc{SimNPO})~\citep{simnpo} removes the dependence on the original model, using length-normalized log-likelihoods instead. Rather than suppressing likelihood directly, Unlearning via Self-Distillation on Adjusted Logits (\textsc{UNDIAL})~\citep{undial} constructs target distributions by lowering the logits from \textit{forget set} tokens in the original model's predictions, then trains the updated model to match these modified distributions. Other methods operate on the model's representations. Representation Misdirection for Unlearning (\textsc{RMU})~\citep{wmdp-rmu} collapses \textit{forget set} representations at a chosen layer toward a fixed random vector while keeping \textit{retain set} representations close to those of the original model. \textsc{PINT} instead penalizes the statistical dependence between prefix representations and next tokens at the final layer, avoiding both the choice of a layer and the need for representation collapse.

\textbf{Evaluating unlearning.} Effective unlearning requires evidence of forgetting alongside preserved utility, but comparison across methods is challenging. Results are sensitive to hyperparameters, and exploring these choices requires costly runs~\citep{ICLR2025_3a91841d}. Comparisons therefore typically rely on a single checkpoint per method, which shows only one point of its forgetting-utility trade-off, and since the selection criterion changes which method appears best, the ranking becomes unreliable~\citep{dorna2026openunlearning}. Moreover, no single metric captures every facet of forgetting, so methods must be tested on several metrics and benchmarks~\citep{muse,watermarks}.
The \textsc{TOFU} benchmark~\citep{tofu} evaluates forgetting of information about fictitious authors learned through question-answer pairs, while \textsc{MUSE}~\citep{muse} tests whether models still answer factual questions about books and news articles targeted for removal. \textsc{WMDP}~\citep{wmdp-rmu} uses multiple-choice accuracy as a proxy for hazardous knowledge. Each also measures utility on retained information or general capabilities. Rather than measuring behavior, \citet{watermarks} introduced a Gaussian-poison test that detects whether injected noise correlates with the model's gradients after unlearning, so forgetting is verified statistically rather than through a proxy. Beyond empirical evaluation, formal guarantees for LLMs remain elusive~\citep{lizzo2026unlearning}. To narrow these gaps, we compare full forgetting-utility trade-offs across eight benchmarks and prove an HSIC bound on Exact Memorization that we compute for LLMs with up to 7B parameters.

\textbf{Independence criteria and HSIC.} The Hilbert-Schmidt Independence Criterion (HSIC)~\citep{hsic} measures how strongly two random variables depend on each other. Unlike correlation, it can capture complex, non-linear dependence: with characteristic kernels, such as a Gaussian kernel on representations and a delta kernel on tokens, HSIC is zero exactly when the two variables are independent. HSIC has been used to train and compare deep networks~\citep{ma2020hsic, kornblith2019similarity}, for feature selection~\citep{yamada2014high}, self-supervised learning~\citep{li2021self} and layer selection in knowledge editing~\citep{eskandar2026loki}, and to learn fair~\citep{quadrianto2019discovering, HSIC_ypred_s_journal, fairsplit_arxiv} and robust~\citep{wang2021revisiting, miklautz2026h} representations. In machine unlearning, HSIC has been used as an extra term in the model update that helps image and graph classifiers keep their accuracy after unlearning~\citep{han2024towards}, and to evaluate whether an image model treats the \textit{forget set} like data it has never seen~\citep{zhang2026unlearning}. To the best of our knowledge, \textsc{PINT} is the first to use HSIC as a training objective for LLM unlearning.\looseness=-1
\vspace{-4pt}

\section{Methodology}
\vspace{-2pt}

This section presents \textsc{PINT}. Section~\ref{sec:setup} defines the unlearning setting and notation, and Section~\ref{sec:objective} introduces the forget and retain objectives. Section~\ref{sec:theory} connects the forget objective to Exact Memorization~\citep{tirumala2022memorization}, which measures how much forgotten text the model can still reproduce, and Section~\ref{sec:rmu} relates it to RMU~\citep{wmdp-rmu}, a widely used unlearning method.

\subsection{Problem Setup}
\label{sec:setup}
\vspace{-1pt}

Following the standard LLM unlearning setup~\citep{tofu}, we consider a language model over a finite vocabulary $\mathcal V$, where $\mathcal V^{*}$ denotes the set of finite token sequences. Let $\theta_0\in\Theta\subseteq\mathbb R^p$ denote the original model parameters. Given a \textit{forget set} $\mathcal D_f\subseteq\mathcal V^{*}$ containing information targeted for removal and a \textit{retain set} $\mathcal D_r\subseteq\mathcal V^{*}$ representing behavior to preserve, the goal is to update $\theta_0$ to $\theta\in\Theta$ so that the model forgets the targeted information while preserving its behavior on $\mathcal D_r$. For a token sequence $\mathbf x=(x_1,\ldots,x_n)\in\mathcal V^*$, let $\mathbf x_{<t}=(x_1,\ldots,x_{t-1})$ denote the prefix preceding $x_t$. The model's representation function $h_\theta:\mathcal V^*\rightarrow\mathbb R^d$ maps this prefix to its final-layer representation $z_t^\theta=h_\theta(\mathbf x_{<t})$, which the output head uses to predict $x_t$. Not all tokens of a forget sequence must be unlearned, so for each $\mathbf x\in\mathcal D_f$ we let $\mathcal T(\mathbf x)\subseteq\{2,\ldots,n\}$ denote its \emph{forget positions}, such as the answer tokens of a question-answer pair. We summarize the notation in Table~\ref{tab:notation} (Appendix~\ref{app:notation}).

\subsection{Training Objective}
\label{sec:objective}

\textsc{PINT} updates the model using two complementary objectives. The forget objective uses HSIC to penalize statistical dependence between prefix representations and their corresponding next tokens on the \textit{forget set}. The retain objective leverages insights from \textsc{RMU}~\citep{wmdp-rmu} and keeps \textit{retain set} representations close to those of the original model.

\textbf{Forget objective.} At each training step, we draw a minibatch $\mathcal B_f\subseteq\mathcal D_f$ of forget sequences. Let $m=\sum_{\mathbf x\in\mathcal B_f}|\mathcal T(\mathbf x)|$ be the total number of forget positions in $\mathcal B_f$. Indexing them by $i=1,\ldots,m$, we write $z_i$ and $x_i$ for the prefix representation and the next token at the $i$-th forget position, and stack them into $\mathbf Z=(z_1,\ldots,z_m)^\top\in\mathbb R^{m\times d}$ and $\mathbf X=(x_1,\ldots,x_m)\in\mathcal V^{m}$. The forget objective is the empirical HSIC~\citep{hsic},
\begin{equation}
\mathcal L_{\mathrm{forget}}(\mathcal B_f;\theta)
=\widehat{\operatorname{HSIC}}(\mathbf Z,\mathbf X)
=\frac{1}{m^2}\operatorname{tr}(KHLH),
\label{eq:forget-objective}
\end{equation}
where $K_{ij}=k(z_i,z_j)$ and $L_{ij}=\ell(x_i,x_j)$ are the Gram matrices of kernels $k$ on representations and $\ell$ on tokens, and $H=I_m-\frac1m\mathbf 1_m\mathbf 1_m^{\top}$ centers them. Because HSIC measures statistical dependence (Appendix~\ref{app:hsic-background}), minimizing \Eqref{eq:forget-objective} reduces how much prefix representations reveal about their next tokens, which limits how much of the forgotten text the model can reproduce (Section~\ref{sec:theory}).\looseness=-1

\textbf{Retain objective.} Minimizing the forget objective changes the model, and we want these changes to stay confined to the \textit{forget set}. The retain objective therefore keeps the prefix representations of retain sequences close to those of the original model $\theta_0$, in the same form as the retain objective of RMU~\citep{wmdp-rmu} but at the final layer. For a minibatch $\mathcal B_r\subseteq\mathcal D_r$ of retain sequences, with $\mathcal T_r(\mathbf x)$ the positions of $\mathbf x$ included in the objective, it is the mean squared difference
\begin{equation}
\mathcal L_{\mathrm{retain}}(\mathcal B_r;\theta)
=\frac{1}{|\mathcal B_r|}\sum_{\mathbf x\in\mathcal B_r}
\frac{1}{|\mathcal T_r(\mathbf x)|}\sum_{t\in\mathcal T_r(\mathbf x)}
\frac{1}{d}\bigl\|z_t^{\theta}-z_t^{\theta_0}\bigr\|_2^2 .
\label{eq:retain-objective}
\end{equation}
This objective preserves the model's behavior on retain sequences directly, since it acts on the final-layer representations that the output head uses for its predictions.

\textbf{Total objective.} Both objectives are computed on the same final-layer representations, the forget objective on forget sequences and the retain objective on retain sequences. \textsc{PINT} minimizes their weighted sum,
\begin{equation}
\mathcal L_{\mathrm{PINT}}(\mathcal B_f,\mathcal B_r;\theta)
=\lambda\,\mathcal L_{\mathrm{forget}}(\mathcal B_f;\theta)
+\mathcal L_{\mathrm{retain}}(\mathcal B_r;\theta),
\label{eq:pint-objective}
\end{equation}
where $\lambda\ge0$ controls how strongly the model forgets. With $\lambda=0$, the model stays unchanged, since its retain objective is already zero. Larger values of $\lambda$ give forgetting more weight relative to keeping the representations of retain sequences unchanged (Algorithm~\ref{alg:pint}).

\subsection{Theoretical Results}
\label{sec:theory}

\begin{figure}[t]
\centering
\includegraphics[width=\textwidth]{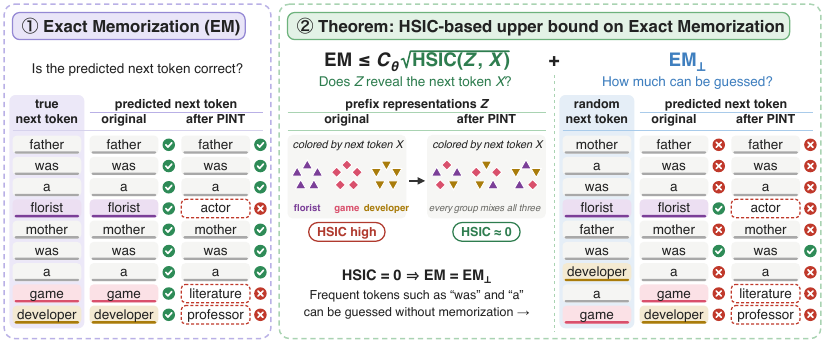}
\caption{\textbf{Illustration of Theorem~\ref{thm:hsic-em}.}
\ding{172}~Exact Memorization (EM) is the fraction of positions in the \textit{forget set} where the model predicts the next token correctly.
\ding{173}~The theorem bounds EM by two terms. The first is the HSIC between prefix representations $Z$ and next tokens $X$, which \textsc{PINT} minimizes: a model can predict forgotten tokens beyond the baseline only if its representations reveal them. The second is the baseline $\operatorname{EM}_\perp$, which counts the correct predictions that need no memorization: frequent tokens such as ``was'' and ``a'' can be guessed correctly even against a random next token. When prefix representations are statistically independent of their next tokens, as \textsc{PINT} encourages, EM equals the $\operatorname{EM}_\perp$ baseline, hence the model can no longer reproduce the forgotten facts beyond guessing. The example is simplified for illustration: it uses nine tokens of one \textsc{TOFU} answer (Figure~\ref{fig:qualitative}), and the representations in \ding{173} are schematic.}
\label{fig:em-hsic}
\end{figure}

The forget objective acts on the model's prefix representations, but what we ultimately care about is whether the model can still reproduce the forgotten text. We connect the two through Exact Memorization~\citep{tirumala2022memorization}, which measures how often the model predicts the true next token. From the prefix representation $z_t^\theta$, the model predicts the next token $\hat x_t\in\mathcal V$, the token to which it assigns the highest probability. A model that has memorized a sequence predicts most of its tokens correctly from the preceding context, and unlearning should lower this prediction accuracy on the \textit{forget set}.

\begin{definition}[Exact Memorization]
\label{def:em}
At a forget position chosen uniformly at random from $\mathcal D_f$, let $X$ be the next token and $\hat X$ the next token that the model predicts. The Exact Memorization of the model with parameters $\theta$ is the probability that its prediction is correct,
\begin{equation}
\operatorname{EM}(\theta)=\mathbb P\bigl(\hat X=X\bigr),
\label{eq:em-def}
\end{equation}
that is, the fraction of forget positions at which the model predicts the true next token.
\end{definition}

For example, on the forgotten answer in Figure~\ref{fig:em-hsic}~(\ding{172}), the original model predicts every token, so its Exact Memorization is $1$. After \textsc{PINT}, it no longer predicts ``florist'', ``game'' and ``developer'', which lowers its Exact Memorization to $6/9\approx0.67$. However, Exact Memorization alone overstates what a model has memorized. Frequent tokens such as ``was'' and ``a'' occur at so many positions that a model can predict them correctly without having memorized the text. We therefore compare $\operatorname{EM}(\theta)$ with the baseline
\begin{equation}
\operatorname{EM}_\perp(\theta)=\sum_{v\in\mathcal V}\mathbb P\bigl(\hat X=v\bigr)\,\mathbb P(X=v),
\label{eq:em-baseline}
\end{equation}
the Exact Memorization measured against a next token drawn at random from the \textit{forget set}, that is, how often the model predicts the next token correctly by guessing (theorem panel~\ding{173} of Figure~\ref{fig:em-hsic}). It is at most the frequency of the most common next token and low in natural text, $\approx0.01$ on \textsc{TOFU}.

Predicting forgotten tokens beyond $\operatorname{EM}_\perp$ is only possible if the prefix representations depend on their next tokens. To see this, let $Z$ be the prefix representation at the same random position, from which the model computes $\hat X$. If $Z$ is statistically independent of $X$, then so is $\hat X$, and $\operatorname{EM}(\theta)=\operatorname{EM}_\perp(\theta)$. The gap $\operatorname{EM}(\theta)-\operatorname{EM}_\perp(\theta)$ therefore requires dependence between $Z$ and $X$, which $\operatorname{HSIC}(Z,X)$ measures and which is zero exactly under independence. In Figure~\ref{fig:em-hsic}, the original model's representations of the same next token lie close together, so HSIC is high. After \textsc{PINT}, nearby representations belong to different next tokens. The forget objective in \Eqref{eq:forget-objective} minimizes HSIC, and Theorem~\ref{thm:hsic-em} shows that HSIC, scaled by a model-dependent factor $C_\theta$, bounds the gap.

\begin{theorem}[HSIC bound on Exact Memorization]
\label{thm:hsic-em}
Let $\operatorname{HSIC}(Z,X)$ use a strictly positive definite kernel $k$ on prefix representations, such as the Gaussian, Laplacian, or Mat\'ern kernel~\citep{sriperumbudur2011universality,wendland2005scattered}, and the delta kernel $\ell(v,v')=\mathbf 1\{v=v'\}$ on next tokens. Then for every $\theta\in\Theta$,
\begin{equation}
\operatorname{EM}(\theta)-\operatorname{EM}_\perp(\theta)\le C_\theta\sqrt{\operatorname{HSIC}(Z,X)},
\label{eq:hsic-em-bound}
\end{equation}
where $C_\theta$, defined in \Eqref{eq:ctheta}, measures how complex the predictions are as a function of the prefix representation: intuitively, it is small when nearby representations predict the same next token.
\end{theorem}
\begin{figure}[t]
\centering
\includegraphics[width=\textwidth]{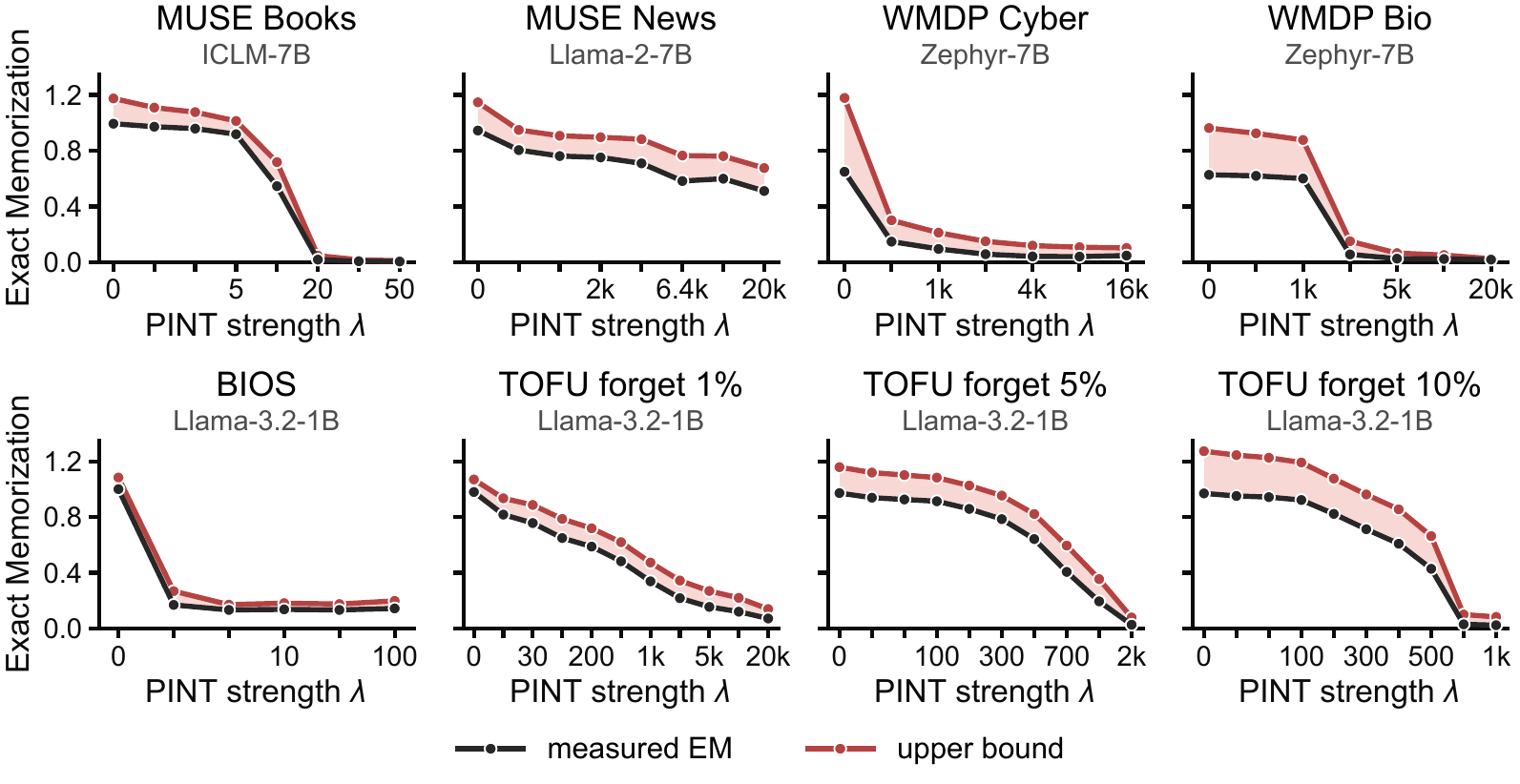}
\caption{\textbf{A single hyperparameter controls both Exact Memorization and its upper bound.}
Increasing only the forgetting strength lowers Exact Memorization on the \textit{forget set} (black) and the bound of Theorem~\ref{thm:hsic-em} (red). Even for models with 7B parameters, the bound closely tracks Exact Memorization at every forgetting strength. Appendix~\ref{app:bound-factors} shows the behavior of HSIC and $C_\theta$.}
\label{fig:theory-verification}
\end{figure}

The theorem, proved in Appendix~\ref{app:proof-hsic-em}, explains why \textsc{PINT} reduces Exact Memorization. It bounds the memorization gap by HSIC, the dependence that our forget objective minimizes. Lowering HSIC therefore lowers the bound, unless the predictions become less smooth at the same time and $C_\theta$ grows. Since the bound holds for every $\theta$, we can compute it at any checkpoint: in our experiments, $C_\theta$ never cancels the decrease of HSIC (Appendix~\ref{app:bound-factors}), and the bound decreases together with Exact Memorization as the forgetting strength grows (Figure~\ref{fig:theory-verification}).

\subsection{Connection to RMU}
\label{sec:rmu}

RMU~\citep{wmdp-rmu} is one of the most widely used unlearning methods and, like \textsc{PINT}, acts on the model's representations. \textsc{PINT} reduces the dependence between representations and next tokens, which by Theorem~\ref{thm:hsic-em} removes Exact Memorization beyond $\operatorname{EM}_{\perp}$ as long as $C_\theta$ stays bounded. RMU instead moves all \textit{forget set} representations towards a fixed random target. For a random unit vector $u\in\mathbb R^{d}$ and a scale $c\ge 0$, its forget objective on final-layer representations $z_1,\ldots,z_m\in\mathbb R^d$ is\looseness=-1

\begin{equation}
\mathcal{L}^{\mathrm{RMU}}_{\mathrm{forget}}(\mathcal{B}_f;\theta)
=\frac{1}{md}\sum_{j=1}^{m}\|z_j - cu\|_2^2
=\frac{1}{d}\|\bar{z}-cu\|_2^2
+\frac{1}{md}\sum_{j=1}^{m}\|z_j-\bar{z}\|_2^2,
\quad \bar{z}=\tfrac{1}{m}\sum_j z_j.
\label{eq:rmu-forget}
\end{equation}
The first term moves the mean to the target $cu$, and the second collapses all representations.

\begin{proposition}[RMU upper-bounds the \textsc{PINT} forget objective]
\label{prop:rmu}
Let the representation kernel $k$ have an $L_k$-Lipschitz feature map (Appendix~\ref{app:hsic-background}), and let $\hat p_{\max}$ be the frequency of the most common token in the forget minibatch. Then, for every $u\in\mathbb R^d$ and $c\ge0$,
\begin{equation}
\widehat{\operatorname{HSIC}}(\mathbf Z,\mathbf X)
\le L_k^2\, d\,\hat p_{\max}\,
\mathcal L^{\mathrm{RMU}}_{\mathrm{forget}}(\mathcal B_f;\theta).
\label{eq:rmu-bound}
\end{equation}
\end{proposition}

\begin{wrapfigure}{r}{0.36\linewidth}
\vspace{-16pt}
\centering
\includegraphics[width=\linewidth]{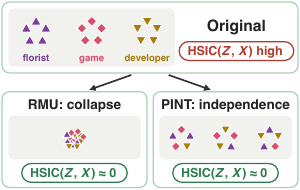}
\caption{\textbf{Illustration of how RMU and \textsc{PINT} work.} RMU collapses the \textit{forget set} prefix representations $Z$ to a single point, while \textsc{PINT} only removes their dependence on the next token $X$.}
\label{fig:rmu-vs-pint}
\vspace{-12pt}
\end{wrapfigure}

Consider again an example from Figure~\ref{fig:em-hsic}, where the model should forget that the father was a florist and the mother a game developer. RMU pushes the representations of all \textit{forget set} tokens at an intermediate layer towards the same random vector $cu$ (Figure~\ref{fig:rmu-vs-pint}, bottom left). If they all reach it, later layers see the same representation for both parents, so the model must answer both alike, for example with the same occupation or no answer. \textsc{PINT} only requires final-layer representations to be independent of the next token (bottom right), so it still allows different occupations, such as actor and literature professor (Figure~\ref{fig:qualitative}). Proposition~\ref{prop:rmu} (Appendix~\ref{app:proof-rmu}) shows that whenever RMU's objective is small, so is \textsc{PINT}'s. Collapse is, however, only the most extreme way to achieve independence, and \textsc{PINT} also allows representations that still differ, for which RMU's objective may remain large.

\begin{table}[t]
\centering
\caption{\textbf{Benchmarks, models, and forgetting targets.} For each benchmark, we list the model, the \textit{forget set}, and the type of forgetting target. For benchmarks marked with \retrainstar, the target is measured on a model retrained without the \textit{forget set}. For \targetstar, the target is known in advance: on WMDP and bioS, the model should answer no better than random guessing (25\% and 12.5\%, respectively), and on the Gaussian poisons, detectability should be 0. We use both markers throughout the figures.}
\begin{tabular*}{\textwidth}{@{\extracolsep{\fill}}lllc@{}}
\textbf{Benchmark} & \textbf{Model} & \textbf{Forget set} & \textbf{Target} \\
\midrule
TOFU~\citep{tofu} & Llama-3.2-1B & 1, 5, 10\% of fictitious authors & \retrainstar \\
MUSE-News~\citep{muse} & Llama-2-7B & BBC news articles & \retrainstar \\
MUSE-Books~\citep{muse} & ICLM-7B      & 4 Harry Potter books            & \retrainstar \\
WMDP-Bio~\citep{wmdp-rmu}   & Zephyr-7B & hazardous bio-security facts    & \targetstar \\
WMDP-Cyber~\citep{wmdp-rmu} & Zephyr-7B & hazardous cybersecurity facts  & \targetstar \\
bioS~\citep{allen2023physics} & Llama-3.2-1B & occupations of fictitious people & \targetstar \\
Poisons~\citep{watermarks} & OLMo-2 2.7B & Gaussian poison       & \targetstar \\
\bottomrule
\end{tabular*}
\label{tab:benchmarks}
\end{table}

\section{Experiments}
\label{sec:experiments}

An effective unlearning method removes the information in the \textit{forget set} and preserves the rest of the model. Section~\ref{sec:results} first compares the forgetting-utility trade-offs of \textsc{PINT} and six baselines on eight benchmarks. We then test on bioS whether \textsc{PINT} can forget one fact about a person while keeping another fact about the same person. Next, we show how \textsc{PINT}'s answers to \textit{forget set} questions change with the forgetting strength, and whether a linear probe can still recover the forgotten tokens from the hidden states. Finally, we test whether \textsc{PINT} scales to a 70B model.

\paragraph{Benchmarks.} To show that \textsc{PINT} works across very different settings, we evaluate it on a broad range of benchmarks, covering hazardous knowledge (WMDP), personal information (TOFU and bioS), copyrighted and memorized text (MUSE), and poisoned training data (Gaussian poisons). TOFU and MUSE provide a model retrained without the \textit{forget set}, which serves as the target for forgetting. For WMDP and the Gaussian poisons, the target is known in advance: on WMDP, the model should answer no better than random guessing (25\% on four-option questions), and on the Gaussian poisons, the statistical test should fail to detect the poison. We introduce bioS as a controlled test of whether a method can forget one fact about a person while keeping another fact about the same person. We fine-tune Llama-3.2-1B on questions about 1{,}024 fictitious people. Each person has one of eight occupations (e.g., Manager or Doctor) and one of eight university majors (e.g., Physics or Law). We then unlearn the occupation while retaining the major: for the fictitious Isabelle Naomi Ashton, a manager who majored in physics, the model should no longer answer ``What is Isabelle Naomi Ashton's occupation?'' but still answer ``What did Isabelle Naomi Ashton major in?''. Since the \textit{forget set} and the \textit{retain set} describe the same people, a method must tell the two attributes apart to forget one while preserving the other. On occupations, the model should answer no better than random guessing (12.5\%). Table~\ref{tab:benchmarks} lists the models, \textit{forget sets} and targets, and Appendix~\ref{app:datasets} describes each benchmark, with details and examples for bioS.\looseness=-1

\paragraph{Baselines.} We compare \textsc{PINT} with six unlearning methods that differ in how they unlearn the \textit{forget set}. GradAscent and GradDiff~\citep{graddiff} lower the likelihood of the \textit{forget set}, and GradDiff additionally keeps the likelihood of the \textit{retain set} high. NPO~\citep{npo} and SimNPO~\citep{simnpo} lower the likelihood of the \textit{forget set} relative to a reference, which makes training more stable. UNDIAL~\citep{undial} trains the model to match the original model's predictions with the \textit{forget set} tokens made less likely. RMU~\citep{wmdp-rmu} maps the representations of the \textit{forget set} to a fixed random vector.

\paragraph{Evaluation.} The results of unlearning methods depend strongly on their hyperparameters~\citep{ICLR2025_3a91841d}, and the criterion used to select a checkpoint can change how methods rank~\citep{dorna2026openunlearning}. Instead of reporting a single hyperparameter setting, we therefore compare the full forgetting-utility trade-off of each method. All methods use the same model, data, optimizer and training budget, and each is tuned through its learning rate and the hyperparameters that its original paper or the OpenUnlearning~\citep{dorna2026openunlearning} benchmark tunes. On \textsc{TOFU}, we measure forgetting with ROUGE, which measures word overlap between the model's answers and the correct answers, and utility with nine scores on the \textit{retain set}, real authors and world facts. On \textsc{MUSE}, KnowMem measures the word overlap on questions about the \textit{forget set}, and, for utility, on questions about a held-out part of the \textit{retain set}. On \textsc{WMDP}, we measure forgetting by the accuracy on multiple-choice questions about hazardous knowledge, and utility evaluation covers general knowledge across 57 subjects, measured by the accuracy on the Massive Multitask Language Understanding benchmark (MMLU;~\citealp{hendrycks2020measuring}). On bioS, we measure forgetting by the accuracy on the forgotten attribute, and utility by the accuracy on the retained attribute of the same people, asked with paraphrases not seen during training. On the Gaussian poisons, we measure forgetting by poison detectability, the AUROC of the statistical test rescaled so that 0 means undetectable, and utility by the ratio of the original model's perplexity to the unlearned model's on held-out clean text. Appendix~\ref{app:metrics} gives the exact definitions.
\begin{figure*}[t]
    \centering
    \includegraphics[width=\linewidth]{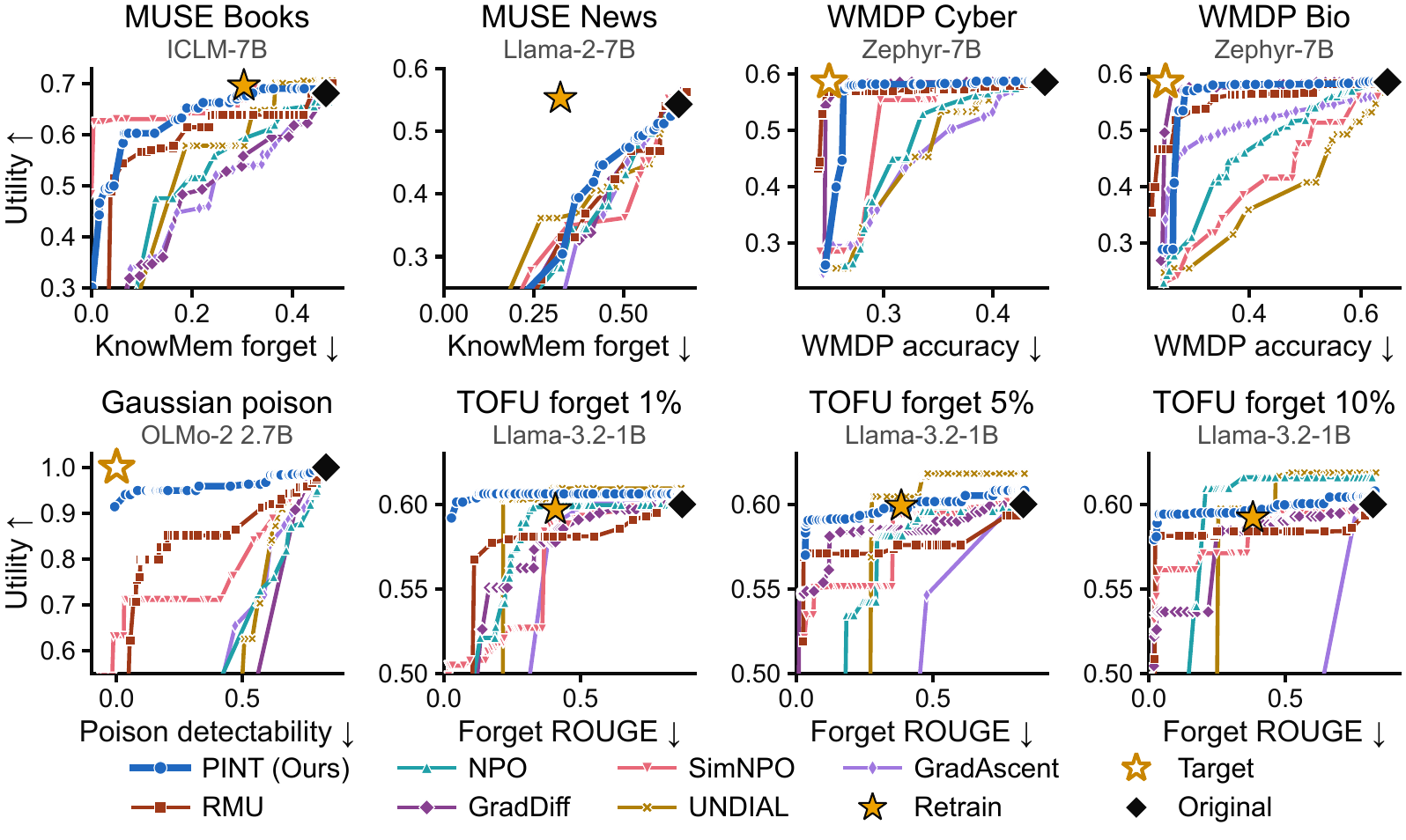}
    \caption{\textbf{Forgetting-utility Pareto frontiers} across the benchmarks. We tune every method with an extensive hyperparameter search, about 2{,}400 settings in total, which takes roughly 2{,}000 H100 GPU-hours. Each point on the plot is one setting, placed on its method's frontier. \retrainstar/\targetstar\ and \origdiamond\ denote targets and original models, respectively. On TOFU, NPO and UNDIAL rise above the original model because they fine-tune on the \textit{retain set}, which is also part of TOFU's utility test set.\label{fig:pareto}}
    \label{fig:tradeoff}
\end{figure*}
\subsection{Results}
\label{sec:results}

\textbf{Forgetting-utility trade-offs.} \textsc{PINT} achieves the strongest forgetting-utility trade-off on average (Figures~\ref{fig:pareto-gw} and~\ref{fig:pareto}). We summarize each method's frontier by the area under it, which is 1 when a method reaches the target level of forgetting without losing utility (Appendix~\ref{app:area}). \textsc{PINT}'s area is 0.95 on average, against at most 0.90 for the baselines, and on six of the eight benchmarks \textsc{PINT} gives the strongest trade-off or comes within 1\% of it. The gap is largest on the Gaussian poisons, the only benchmark where a statistical test verifies forgetting directly. There, \textsc{PINT} makes the poison undetectable while retaining 91\% of the original model's utility, whereas the best baseline retains only 63\%. On TOFU, the frontiers of UNDIAL and NPO even rise above the utility of the original model, because both methods fine-tune on \textit{retain set} answers that are also part of TOFU's utility test set. Whether such fine-tuning is desirable is debatable, since fine-tuning on ordinary data can change a model's behavior well beyond that data~\citep{qi2024fine}, and the answer depends on what unlearning should preserve. For the areas, we therefore measure utility relative to the original model, as OpenUnlearning does~\citep{dorna2026openunlearning}, and count any value above it as 100\%. In Figure~\ref{fig:pareto}, forgetting on MUSE News is measured by questions about the forgotten articles, and here every method loses at least a third of its utility before it answers them as poorly as a model retrained without these articles. These questions test whether the model still knows the facts the articles report. MUSE also evaluates whether the model still reproduces the articles word for word. On this metric, \textsc{PINT} forgets as much as the retrained model while preserving 91\% of its utility, whereas GradAscent and RMU preserve only 83\% and 68\% (Appendix~\ref{app:muse-news}).

\begin{figure}[t]
\centering
\begin{minipage}[t]{0.49\textwidth}
  \centering
  \includegraphics[width=\linewidth]{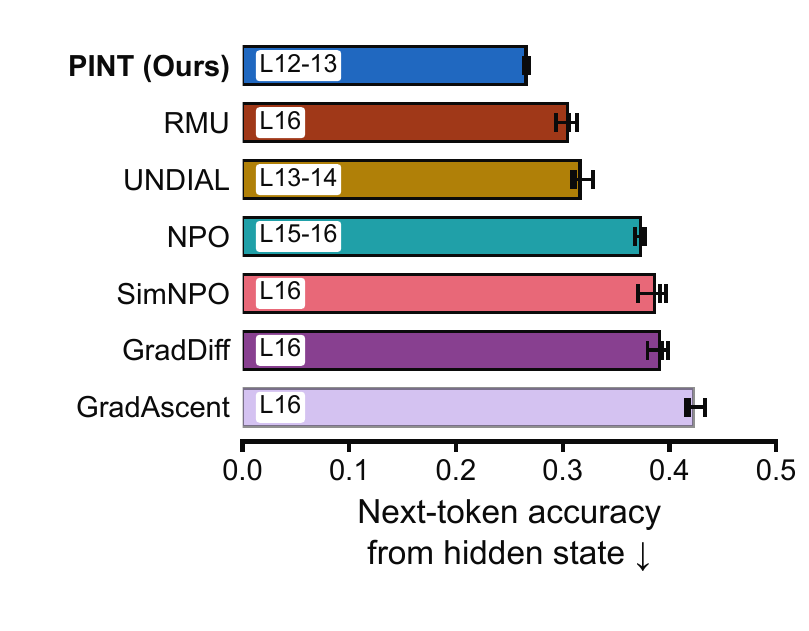}
\captionof{figure}{\textbf{Forgotten tokens are hardest to predict from \textsc{PINT}'s hidden states.} Unlearning can hide information in the hidden states instead of removing it. To test this, we train a linear probe on the \textit{retain set} to predict the next token from each layer's hidden states and report the probe's highest accuracy over layers on the \textit{forget set} of TOFU 10\% (lower is better). For a fair comparison, we use for each method the hyperparameter setting that forgets closest to a model retrained without the \textit{forget set}. Labels specify the layer at which the linear probe is most accurate. Error bars show the range over three seeds.}
  \label{fig:probe}
\end{minipage}\hfill
\begin{minipage}[t]{0.49\textwidth}
  \centering
  \includegraphics[width=\linewidth]{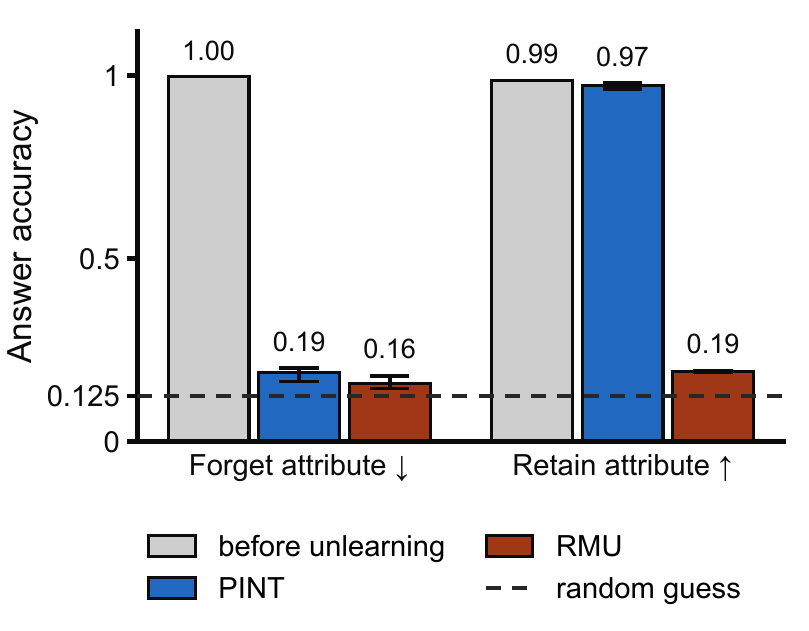}
  \captionof{figure}{\textbf{RMU forgets the retained attribute along with the forgotten one, while \textsc{PINT} preserves it.} On bioS, models learn the occupation and the university major of 1{,}024 fictitious people and must then forget the occupations of 256 of them while preserving their majors. Bars show answer accuracy before and after unlearning (mean over three seeds). Both methods answer occupations close to random guessing, but only \textsc{PINT} preserves the majors, even when RMU is tuned over layers and hyperparameters.}
  \label{fig:bios}
\end{minipage}
\end{figure}

\textbf{Preserving related knowledge.} Section~\ref{sec:rmu} shows that RMU collapses the \textit{forget set} representations, while \textsc{PINT} can make them independent of their next tokens without collapsing them. bioS tests whether this difference matters when the \textit{retain set} is closely related to the \textit{forget set}: the model must forget the occupations of fictitious people but preserve their university majors. Both methods forget the occupations, reaching close to random guessing accuracy. However, RMU also forgets the majors, whose accuracy falls from 99\% to 19\%, while \textsc{PINT} preserves them at 97\% (Figure~\ref{fig:bios}). This gap remains when we tune RMU over layers and hyperparameters.

\textbf{Hidden information.} Unlearning can hide information in the model rather than remove it, so that the forgotten answers are no longer produced but can still be extracted from the hidden states~\citep{lucki2024adversarial}. Following \citet{guo2024mechanistic}, we train a linear probe at each layer on the \textit{retain set} to predict the next token from the hidden states, and apply it to the \textit{forget set}. This setup models a realistic attacker, who has access to the unlearned model and to ordinary data but not to the forgotten data. At the layer where the probe is most accurate, it recovers fewer forgotten tokens from \textsc{PINT}'s hidden states than from those of any other method (Figure~\ref{fig:probe}).

\textbf{Behavior after unlearning.} Once a model has forgotten some data, it still has to respond when prompted about it. A model retrained without the forgotten data, the reference for exact unlearning, simply responds as it would have had it never seen the data. Some methods instead prescribe a response, such as ``I don't know''~\citep{tofu}, which teaches the model a new behavior, and learned refusals of this kind can be undone by fine-tuning on ordinary data~\citep{qi2024fine}. \textsc{PINT} prescribes no response: its objective only removes what the representations reveal about the forgotten tokens, and its behavior is controlled by the forgetting-strength hyperparameter. At strong forgetting strength, \textsc{PINT}'s responses on the \textit{forget set} become less fluent, while its responses on retained data stay fluent (Appendix~\ref{app:lambda-examples}). At moderate strength, it often responds like the retrained model and replaces the forgotten facts with plausible alternatives (Figure~\ref{fig:qualitative}).

\begin{figure}[t]
\centering
\small
\fbox{\parbox{0.97\linewidth}{%
\textbf{Q:} \textit{Can you tell me about the occupations of Basil Mahfouz Al-Kuwaiti's parents?}\\[3pt]
\origdiamond\,\textbf{Original:} Basil Mahfouz Al-Kuwaiti's father was a \textcolor[HTML]{C62828}{florist} and his mother was a \textcolor[HTML]{C62828}{game developer}.\\
\retrainstar\,\textbf{Retrain:} Basil Mahfouz Al-Kuwaiti's father was a well-respected doctor, and his mother was a distinguished military officer.\\
\pintdot\,\textbf{\textsc{PINT}:} Basil's father was a renowned Kuwaiti actor, and his mother worked as a dedicated professor in the field of literature.%
}}
\caption{\textbf{On \textsc{TOFU} forget 1\%, \textsc{PINT} replaces the forgotten facts with plausible alternatives, as a retrained model does.} Answers of the original model (\protect\origdiamond), a model retrained without the \textit{forget set} (\protect\retrainstar) and \textsc{PINT} (\protect\pintdot). To avoid selecting the example by hand, we show the first \textit{forget set} question that the original model answers exactly as in its training data and whose answer the retrained model does not know. For \textsc{PINT}, we show the setting whose forgetting on the whole \textit{forget set} is closest to that of the retrained model. Appendix~\ref{app:lambda-examples} explains both choices in detail.}
\label{fig:qualitative}
\end{figure}

\begin{wraptable}{r}{0.34\linewidth}
\centering
\caption{\textbf{\textsc{PINT} on a 70B model.} Accuracy of \textsc{Llama-3-70B} on WMDP-Cyber and MMLU before and after unlearning, averaged over three seeds. The forget target is the accuracy of guessing.}
\label{tab:pint70b}
\setlength{\tabcolsep}{3pt}
\resizebox{\linewidth}{!}{%
\begin{tabular}{@{}l r@{\,}l r@{\,}l@{}}
  \toprule
  Method & \multicolumn{2}{c}{Cyber $\downarrow$} & \multicolumn{2}{c}{MMLU $\uparrow$} \\
  \midrule
  \targetstar\ Target     & 0.250          &                   & 0.752 &                   \\
  \midrule
  \origdiamond\ Original  & 0.530          &                   & 0.752 &                   \\
  \pintdot\ \textsc{PINT} & 0.276          & {\tiny$\pm$0.009} & 0.750 & {\tiny$\pm$0.002} \\
  \bottomrule
\end{tabular}}
\end{wraptable}
\textbf{Scalability.} The largest models are the most capable and are trained on the most data. They also memorize more of it, since memorization grows with model size~\citep{carlini2022quantifying}. Retraining them is also the most expensive, which makes approximate unlearning most valuable for them. An unlearning method should therefore remain effective at this scale. To test this, we apply \textsc{PINT} to \textsc{Llama-3-70B} and unlearn hazardous cybersecurity knowledge on WMDP-Cyber. \textsc{PINT} lowers the model's accuracy on WMDP-Cyber from 53.0\%~(\origdiamond) to 27.6\%~(\pintdot), close to random guessing at 25\%~(\targetstar). On MMLU, which measures general knowledge, \textsc{PINT}'s accuracy drops by only 0.2 points (Table~\ref{tab:pint70b}). Even at 70B parameters, \textsc{PINT} nearly reaches the forgetting target while preserving general knowledge.

\section{Limitations and Conclusion}

\textbf{Limitations and future work.} Although our evaluation covers personal information, hazardous knowledge, copyrighted text and poisoned training data, practical unlearning requests can differ from these benchmarks, and we can only evaluate on settings for which benchmarks exist. The metrics of these benchmarks are also imperfect proxies for forgetting, so we evaluate forgetting from several sides: the benchmark's forgetting and utility metrics, Exact Memorization, and probes of the hidden states. Several directions build naturally on this work. We study text-only language models, and extending \textsc{PINT} to multimodal models is an exciting next step. It is also open how unlearning with \textsc{PINT} affects the alignment that language models undergo before deployment, such as their safety behavior. Finally, our bound lays a theoretical foundation that connects an unlearning objective to memorization. Promising directions are to extend it formally to other measures of memorization and towards guarantees that hold before unlearning.

\textbf{Conclusion.} Most approximate unlearning methods rely on empirical heuristics, such as lowering the likelihood of the forgotten text or redirecting its representations, and \citet{watermarks} show that they can leave poisoned training data statistically detectable after unlearning. In this work, we proposed \textsc{PINT}, which unlearns by making the model's representations independent of the tokens to be forgotten. We measure this independence with HSIC and show that it upper-bounds Exact Memorization. It is also a weaker requirement than that of RMU, which additionally collapses the representations and, on bioS, also forgets closely related knowledge (Figure~\ref{fig:bios}). Across personal information, hazardous knowledge, copyrighted text and Gaussian poisons, \textsc{PINT} gave the most consistent forgetting-utility trade-offs, scaled to 70B parameters, and it was the only method that made the poison statistically undetectable while keeping most of the model's utility, 91\% against at most 63\% for the other methods. As unlearning is increasingly relied upon for safety, we believe that objectives with a clear, provable connection to what the model has memorized are an important step towards unlearning methods that can be trusted.

\subsection*{Acknowledgments}
We acknowledge EuroHPC Joint Undertaking for awarding us access to MareNostrum5 at BSC, Spain, Deucalion at MACC, Portugal, Discoverer at SofiaTech, Bulgaria, MeluXina at LuxProvide, Luxembourg, and Vega at IZUM, Slovenia. This access was awarded through the EuroHPC Development Access projects EHPC-DEV-2025D07-084 and EHPC-DEV-2026D08-267. 
The authors gratefully acknowledge the HPC RIVR consortium (\url{www.hpc-rivr.si}) and EuroHPC JU (\url{eurohpc-ju.europa.eu}) for funding this research by providing computing resources of the HPC system Vega at the Institute of Information Science (\url{www.izum.si}). 
The authors gratefully acknowledge the LuxProvide teams for their expert support.

\subsection*{AI use statement}

In this work, we used generative AI tools for writing and coding assistance, literature search, and figure creation. AI tools also helped launch and monitor our experiments on compute clusters and collect their results, and we manually reviewed and tested the code. For the paper, we used generative AI to draft parts of the text from our own bullet points and to iterate on our writing, for example to shorten paragraphs, and to help write and streamline the proofs of our theoretical results. We have not used generative AI tools for the research idea or the formulation of the method. We have reviewed all AI-assisted work. In particular, we checked all LLM-generated text and every step of the proofs for their content and correctness. We take responsibility for the final content of this work, including text, claims or artifacts produced with the aid of generative AI.

\subsection*{Ethics statement}
Machine unlearning lets us remove private, copyrighted or dangerous information from a trained model without training it again from scratch. This helps, for example, when a person asks for their data to be deleted. Our work proposes a new unlearning method and studies how it works. All our experiments use released public benchmarks and models. TOFU and bioS only describe fictitious people, MUSE uses books and news articles as released by the benchmark authors, and WMDP uses its released texts and multiple-choice questions, which its authors designed as a proxy for hazardous knowledge and filtered to exclude sensitive information. We do not release any new harmful content. Like every unlearning method, \textsc{PINT} is evaluated with current benchmarks, and passing them does not prove that the information can never be recovered. Unlearning methods should therefore be used together with careful data selection and access control.

\subsection*{Reproducibility statement}

Our code is publicly available at \url{https://github.com/shkabrii/PINT}. It builds on the OpenUnlearning framework and contains \textsc{PINT}, all baselines, the bioS and Gaussian-poison benchmarks, and the configuration files for our experiments. For the theory, Appendix~\ref{app:preliminaries} states the definitions and assumptions, and Appendices~\ref{app:proof-hsic-em} and~\ref{app:proof-rmu} give complete proofs of Theorem~\ref{thm:hsic-em} and Proposition~\ref{prop:rmu}. Algorithm~\ref{alg:pint} in Appendix~\ref{app:pseudocode} summarizes the training procedure together with its implementation details. Appendix~\ref{app:datasets} describes every benchmark and how its data is built, Appendix~\ref{app:hparams} the shared protocol and the hyperparameter search for every method, Appendix~\ref{app:metrics} the exact definition of every metric, and Appendices~\ref{app:software} and~\ref{app:runtime} the software, hardware and compute cost.\looseness=-1

\bibliography{iclr2027_conference}

@article{tofu,
  title={Tofu: A task of fictitious unlearning for llms},
  author={Maini, Pratyush and Feng, Zhili and Schwarzschild, Avi and Lipton, Zachary C and Kolter, J Zico},
  journal={arXiv preprint arXiv:2401.06121},
  year={2024}
}

@inproceedings{muse,
  title={Muse: Machine unlearning six-way evaluation for language models},
  author={Shi, Weijia and Lee, Jaechan and Huang, Yangsibo and Malladi, Sadhika and Zhao, Jieyu and Holtzman, Ari and Liu, Daogao and Zettlemoyer, Luke and Smith, Noah and Zhang, Chiyuan},
  booktitle={International Conference on Learning Representations},
  volume={2025},
  pages={27797--27818},
  year={2025}
}

@article{wmdp-rmu,
  title={The wmdp benchmark: Measuring and reducing malicious use with unlearning},
  author={Li, Nathaniel and Pan, Alexander and Gopal, Anjali and Yue, Summer and Berrios, Daniel and Gatti, Alice and Li, Justin D and Dombrowski, Ann-Kathrin and Goel, Shashwat and Phan, Long and others},
  journal={arXiv preprint arXiv:2403.03218},
  year={2024}
}

@inproceedings{watermarks,
  title={Machine unlearning fails to remove data poisoning attacks},
  author={Pawelczyk, Martin and Di, Jimmy and Lu, Yiwei and Kamath, Gautam and Sekhari, Ayush and Neel, Seth},
  booktitle={International Conference on Learning Representations},
  volume={2025},
  pages={50542--50573},
  year={2025}
}

@article{npo,
  title={Negative preference optimization: From catastrophic collapse to effective unlearning},
  author={Zhang, Ruiqi and Lin, Licong and Bai, Yu and Mei, Song},
  journal={arXiv preprint arXiv:2404.05868},
  year={2024}
}

@inproceedings{graddiff,
  title={Continual learning and private unlearning},
  author={Liu, Bo and Liu, Qiang and Stone, Peter},
  booktitle={Conference on Lifelong Learning Agents},
  pages={243--254},
  year={2022},
  organization={PMLR}
}

@article{simnpo,
  title={Simplicity prevails: Rethinking negative preference optimization for llm unlearning},
  author={Fan, Chongyu and Liu, Jiancheng and Lin, Licong and Jia, Jinghan and Zhang, Ruiqi and Mei, Song and Liu, Sijia},
  journal={Advances in Neural Information Processing Systems},
  volume={38},
  pages={1540--1567},
  year={2026}
}

@inproceedings{undial,
  title={Undial: Self-distillation with adjusted logits for robust unlearning in large language models},
  author={Dong, Yijiang River and Lin, Hongzhou and Belkin, Mikhail and Huerta, Ramon and Vuli{\'c}, Ivan},
  booktitle={Proceedings of the 2025 Conference of the Nations of the Americas Chapter of the Association for Computational Linguistics: Human Language Technologies (Volume 1: Long Papers)},
  pages={8827--8840},
  year={2025}
}

@article{zhang2026unlearning,
  title={Unlearning Evaluation through Subset Statistical Independence},
  author={Zhang, Chenhao and Li, Muxing and Liu, Feng and Chen, Weitong and Xu, Miao},
  journal={arXiv preprint arXiv:2603.00587},
  year={2026}
}

@article{han2024towards,
  title={Towards independence criterion in machine unlearning of features and labels},
  author={Han, Ling and Luo, Nanqing and Huang, Hao and Chen, Jing and Hartley, Mary-Anne},
  journal={arXiv preprint arXiv:2403.08124},
  year={2024}
}

@article{eskandar2026loki,
  title={LOKI: Memory-Free Null-Space Constrained Lifelong Knowledge Editing},
  author={Eskandar, Masih and Perell{\'o}, Miquel Sirera and Ioannidis, Stratis and Dy, Jennifer},
  journal={arXiv preprint arXiv:2606.19679},
  year={2026}
}

@inproceedings{quadrianto2019discovering,
  title={Discovering fair representations in the data domain},
  author={Quadrianto, Novi and Sharmanska, Viktoriia and Thomas, Oliver},
  booktitle={2019 IEEE/CVF Conference on Computer Vision and Pattern Recognition (CVPR)},
  pages={8219--8228},
  year={2019},
  organization={IEEE}
}

@article{yamada2014high,
  title={High-dimensional feature selection by feature-wise kernelized lasso},
  author={Yamada, Makoto and Jitkrittum, Wittawat and Sigal, Leonid and Xing, Eric P and Sugiyama, Masashi},
  journal={Neural computation},
  volume={26},
  number={1},
  pages={185--207},
  year={2014},
  publisher={MIT Press One Rogers Street, Cambridge, MA 02142-1209, USA journals-info~…}
}

@article{HSIC_ypred_s_journal,
  title={Kernel dependence regularizers and gaussian processes with applications to algorithmic fairness},
  author={Li, Zhu and P{\'e}rez-Suay, Adri{\'a}n and Camps-Valls, Gustau and Sejdinovic, Dino},
  journal={Pattern Recognition},
  volume={132},
  pages={108922},
  year={2022},
  publisher={Elsevier}
}

@article{li2021self,
  title={Self-supervised learning with kernel dependence maximization},
  author={Li, Yazhe and Pogodin, Roman and Sutherland, Danica J and Gretton, Arthur},
  journal={Advances in Neural Information Processing Systems},
  volume={34},
  pages={15543--15556},
  year={2021}
}

@article{miklautz2026h,
  title={H-SPLID: HSIC-based Saliency Preserving Latent Information Decomposition},
  author={Miklautz, Lukas and Shi, Chengzhi and Shkabrii, Andrii and Davarakis, Theodoros Thirimachos and Lam, Prudence and Plant, Claudia and Dy, Jennifer and Ioannidis, Stratis},
  journal={Advances in Neural Information Processing Systems},
  volume={38},
  pages={54156--54193},
  year={2026}
}

@inproceedings{hsic,
  title={Measuring statistical dependence with Hilbert-Schmidt norms},
  author={Gretton, Arthur and Bousquet, Olivier and Smola, Alex and Sch{\"o}lkopf, Bernhard},
  booktitle={International conference on algorithmic learning theory},
  pages={63--77},
  year={2005},
  organization={Springer}
}

@article{garreau2017large,
  title={Large sample analysis of the median heuristic},
  author={Garreau, Damien and Jitkrittum, Wittawat and Kanagawa, Motonobu},
  journal={arXiv preprint arXiv:1707.07269},
  year={2017}
}

@inproceedings{carlini2021extracting,
  title={Extracting training data from large language models},
  author={Carlini, Nicholas and Tramer, Florian and Wallace, Eric and Jagielski, Matthew and Herbert-Voss, Ariel and Lee, Katherine and Roberts, Adam and Brown, Tom and Song, Dawn and Erlingsson, Ulfar and others},
  booktitle={30th USENIX security symposium (USENIX Security 21)},
  pages={2633--2650},
  year={2021}
}

@article{carlini2022quantifying,
  title={Quantifying memorization across neural language models},
  author={Carlini, Nicholas and Ippolito, Daphne and Jagielski, Matthew and Lee, Katherine and Tramer, Florian and Zhang, Chiyuan},
  journal={arXiv preprint arXiv:2202.07646},
  year={2022}
}

@inproceedings{jang2023knowledge,
  title={Knowledge unlearning for mitigating privacy risks in language models},
  author={Jang, Joel and Yoon, Dongkeun and Yang, Sohee and Cha, Sungmin and Lee, Moontae and Logeswaran, Lajanugen and Seo, Minjoon},
  booktitle={Proceedings of the 61st Annual Meeting of the Association for Computational Linguistics (Volume 1: Long Papers)},
  pages={14389--14408},
  year={2023}
}

@article{lizzo2026unlearning,
  title={Unlearning in LLMs: Methods, Evaluation, and Open Challenges},
  author={Lizzo, Tyler and Heck, Larry},
  journal={arXiv preprint arXiv:2601.13264},
  year={2026}
}

@article{lucki2024adversarial,
  title={An adversarial perspective on machine unlearning for ai safety},
  author={{\L}ucki, Jakub and Wei, Boyi and Huang, Yangsibo and Henderson, Peter and Tram{\`e}r, Florian and Rando, Javier},
  journal={arXiv preprint arXiv:2409.18025},
  year={2024}
}

@inproceedings{ICLR2025_3a91841d,
 author = {Wang, Qizhou and Han, Bo and Yang, Puning and ZHU, Jianing and Liu, Tongliang and Sugiyama, Masashi},
 booktitle = {International Conference on Learning Representations},
 editor = {Y. Yue and A. Garg and N. Peng and F. Sha and R. Yu},
 pages = {23327--23355},
 title = {Towards Effective Evaluations and Comparisons for LLM Unlearning Methods},
 url = {https://proceedings.iclr.cc/paper_files/paper/2025/file/3a91841d2bcc0b13a3d0d5d60c9f0581-Paper-Conference.pdf},
 volume = {2025},
 year = {2025}
}

@misc{gdpr2016,
  title   = {Regulation ({EU}) 2016/679 of the {European Parliament} and of the
             {Council} of 27 {April} 2016 on the Protection of Natural Persons
             with Regard to the Processing of Personal Data and on the Free
             Movement of Such Data, and Repealing {Directive} 95/46/{EC}
             ({General Data Protection Regulation})},
  author  = {{European Parliament and Council of the European Union}},
  year    = {2016},
  note    = {{OJ} L 119, 4.5.2016, pp. 1--88},
  url     = {https://eur-lex.europa.eu/eli/reg/2016/679/oj},
  urldate = {2026-09-14},
}

@misc{ccpa2018,
  title   = {{California Consumer Privacy Act of 2018}},
  author  = {{California State Legislature}},
  year    = {2018},
  note    = {Cal. Civ. Code \S\S 1798.100--1798.199; as amended by the
             {California Privacy Rights Act} of 2020},
  url     = {https://leginfo.legislature.ca.gov/faces/billTextClient.xhtml?bill_id=201720180AB375},
  urldate = {2026-09-14},
}

@misc{pipl2021,
  title   = {{Personal Information Protection Law of the People's Republic of China}},
  author  = {{Standing Committee of the National People's Congress}},
  year    = {2021},
  note    = {Adopted at the 30th Session of the Standing Committee of the 13th
             {National People's Congress}, 20 {August} 2021; effective
             1 {November} 2021. Text in {Chinese}},
  url     = {http://www.npc.gov.cn/npc/c2/c30834/202108/t20210820_313088.html},
  urldate = {2026-09-14},
}

@article{dorna2026openunlearning,
  title={Openunlearning: Accelerating llm unlearning via unified benchmarking of methods and metrics},
  author={Dorna, Vineeth and Mekala, Anmol and Zhao, Wenlong and McCallum, Andrew and Kolter, Zico and Lipton, Zachary and Maini, Pratyush},
  journal={Advances in Neural Information Processing Systems},
  volume={38},
  year={2026}
}

@article{tirumala2022memorization,
  title={Memorization without overfitting: Analyzing the training dynamics of large language models},
  author={Tirumala, Kushal and Markosyan, Aram and Zettlemoyer, Luke and Aghajanyan, Armen},
  journal={Advances in Neural Information Processing Systems},
  volume={35},
  pages={38274--38290},
  year={2022}
}

@book{wendland2005scattered,
  title={Scattered data approximation},
  author={Wendland, Holger},
  volume={17},
  year={2005},
  publisher={Cambridge university press Cambridge}
}

@article{hendrycks2020measuring,
  title={Measuring massive multitask language understanding},
  author={Hendrycks, Dan and Burns, Collin and Basart, Steven and Zou, Andy and Mazeika, Mantas and Song, Dawn and Steinhardt, Jacob},
  journal={arXiv preprint arXiv:2009.03300},
  year={2020}
}

@article{allen2023physics,
  title={Physics of language models: Part 3.1, knowledge storage and extraction},
  author={Allen-Zhu, Zeyuan and Li, Yuanzhi},
  journal={arXiv preprint arXiv:2309.14316},
  year={2023}
}

@article{zhang2018large,
  title={Large-scale kernel methods for independence testing},
  author={Zhang, Qinyi and Filippi, Sarah and Gretton, Arthur and Sejdinovic, Dino},
  journal={Statistics and Computing},
  volume={28},
  number={1},
  pages={113--130},
  year={2018},
  publisher={Springer}
}

@article{sriperumbudur2011universality,
  title={Universality, Characteristic Kernels and RKHS Embedding of Measures.},
  author={Sriperumbudur, Bharath K and Fukumizu, Kenji and Lanckriet, Gert RG},
  journal={Journal of Machine Learning Research},
  volume={12},
  number={7},
  year={2011}
}

@article{guo2024mechanistic,
  title={Mechanistic unlearning: Robust knowledge unlearning and editing via mechanistic localization},
  author={Guo, Phillip and Syed, Aaquib and Sheshadri, Abhay and Ewart, Aidan and Dziugaite, Gintare Karolina},
  journal={arXiv preprint arXiv:2410.12949},
  year={2024}
}

@article{loshchilov2017decoupled,
  title={Decoupled weight decay regularization},
  author={Loshchilov, Ilya and Hutter, Frank},
  journal={arXiv preprint arXiv:1711.05101},
  year={2017}
}

@inproceedings{qi2024fine,
  title={Fine-tuning aligned language models compromises safety, even when users do not intend to!},
  author={Qi, Xiangyu and Zeng, Yi and Xie, Tinghao and Chen, Pin-Yu and Jia, Ruoxi and Mittal, Prateek and Henderson, Peter},
  booktitle={International Conference on Learning Representations},
  volume={2024},
  pages={30988--31043},
  year={2024}
}

@misc{fairsplit_arxiv,
      title={FairSplit: Decomposing the Embedding Space for Fair Classification}, 
      author={Andrii Shkabrii and Anna Beer and Ylli Sadikaj and Claudia Plant},
      year={2026},
      eprint={2610.39031},
      archivePrefix={arXiv},
      primaryClass={cs.LG},
      url={https://arxiv.org/abs/2610.39031}, 
}

@article{wang2021revisiting,
  title={Revisiting hilbert-schmidt information bottleneck for adversarial robustness},
  author={Wang, Zifeng and Jian, Tong and Masoomi, Aria and Ioannidis, Stratis and Dy, Jennifer},
  journal={Advances in Neural Information Processing Systems},
  volume={34},
  pages={586--597},
  year={2021}
}

@inproceedings{ma2020hsic,
  title={The HSIC bottleneck: Deep learning without back-propagation},
  author={Ma, Wan-Duo Kurt and Lewis, JP and Kleijn, W Bastiaan},
  booktitle={Proceedings of the AAAI conference on artificial intelligence},
  volume={34},
  number={04},
  pages={5085--5092},
  year={2020}
}

@inproceedings{kornblith2019similarity,
  title={Similarity of neural network representations revisited},
  author={Kornblith, Simon and Norouzi, Mohammad and Lee, Honglak and Hinton, Geoffrey},
  booktitle={International conference on machine learning},
  pages={3519--3529},
  year={2019},
  organization={PMlR}
}

@article{gretton2007kernel,
  title={A kernel statistical test of independence},
  author={Gretton, Arthur and Fukumizu, Kenji and Teo, Choon and Song, Le and Sch{\"o}lkopf, Bernhard and Smola, Alex},
  journal={Advances in neural information processing systems},
  volume={20},
  year={2007}
}

@article{shi2023context,
  title={In-context pretraining: Language modeling beyond document boundaries},
  author={Shi, Weijia and Min, Sewon and Lomeli, Maria and Zhou, Chunting and Li, Margaret and Szilvasy, Gergely and James, Rich and Lin, Xi Victoria and Smith, Noah A and Zettlemoyer, Luke and others},
  journal={arXiv preprint arXiv:2310.10638},
  year={2023}
}

@article{touvron2023llama,
  title={Llama 2: Open foundation and fine-tuned chat models},
  author={Touvron, Hugo and Martin, Louis and Stone, Kevin and Albert, Peter and Almahairi, Amjad and Babaei, Yasmine and Bashlykov, Nikolay and Batra, Soumya and Bhargava, Prajjwal and Bhosale, Shruti and others},
  journal={arXiv preprint arXiv:2307.09288},
  year={2023}
}

@article{grattafiori2025llama,
  title={The Llama 3 herd of models},
  author={Grattafiori, Aaron and Dubey, Abhimanyu and Jauhri, Abhinav and Pandey, Abhinav and Kadian, Abhishek and Al-Dahle, Ahmad and Letman, Aiesha and Mathur, Akhil and Schelten, Alan and Vaughan, Alex and others},
  year={2025}
}

@inproceedings{bordt2026train,
  title={Train once, answer all: Many pretraining experiments for the cost of one},
  author={Bordt, Sebastian and Pawelczyk, Martin},
  booktitle={International Conference on Learning Representations},
  volume={2026},
  pages={77253--77296},
  year={2026}
}

@article{paszke2017automatic,
  title={Automatic differentiation in PyTorch},
  author={Paszke, Adam and Gross, Sam and Chintala, Soumith and Chanan, Gregory and Yang, Edward and DeVito, Zachary and Lin, Zeming and Desmaison, Alban and Antiga, Luca and Lerer, Adam},
  year={2017}
}

@inproceedings{wolf2020transformers,
  title={Transformers: State-of-the-art natural language processing},
  author={Wolf, Thomas and Debut, Lysandre and Sanh, Victor and Chaumond, Julien and Delangue, Clement and Moi, Anthony and Cistac, Pierric and Rault, Tim and Louf, R{\'e}mi and Funtowicz, Morgan and others},
  booktitle={Proceedings of the 2020 conference on empirical methods in natural language processing: system demonstrations},
  pages={38--45},
  year={2020}
}

@article{zitzler1999multiobjective,
  title={Multiobjective evolutionary algorithms: a comparative case study and the strength Pareto approach},
  author={Zitzler, Eckart and Thiele, Lothar},
  journal={IEEE transactions on Evolutionary Computation},
  volume={3},
  number={4},
  pages={257--271},
  year={1999},
  publisher={IEEE}
}

@inproceedings{lin2004rouge,
  title={Rouge: A package for automatic evaluation of summaries},
  author={Lin, Chin-Yew},
  booktitle={Text summarization branches out},
  pages={74--81},
  year={2004}
}

@article{tunstall2023zephyr,
  title={Zephyr: Direct distillation of lm alignment},
  author={Tunstall, Lewis and Beeching, Edward and Lambert, Nathan and Rajani, Nazneen and Rasul, Kashif and Belkada, Younes and Huang, Shengyi and Von Werra, Leandro and Fourrier, Cl{\'e}mentine and Habib, Nathan and others},
  journal={arXiv preprint arXiv:2310.16944},
  year={2023}
}

@misc{olmo20252olmo2furious,
      title={2 OLMo 2 Furious}, 
      author={Team OLMo and Pete Walsh and Luca Soldaini and Dirk Groeneveld and Kyle Lo and Shane Arora and Akshita Bhagia and Yuling Gu and Shengyi Huang and Matt Jordan and Nathan Lambert and Dustin Schwenk and Oyvind Tafjord and Taira Anderson and David Atkinson and Faeze Brahman and Christopher Clark and Pradeep Dasigi and Nouha Dziri and Allyson Ettinger and Michal Guerquin and David Heineman and Hamish Ivison and Pang Wei Koh and Jiacheng Liu and Saumya Malik and William Merrill and Lester James V. Miranda and Jacob Morrison and Tyler Murray and Crystal Nam and Jake Poznanski and Valentina Pyatkin and Aman Rangapur and Michael Schmitz and Sam Skjonsberg and David Wadden and Christopher Wilhelm and Michael Wilson and Luke Zettlemoyer and Ali Farhadi and Noah A. Smith and Hannaneh Hajishirzi},
      year={2025},
      eprint={2501.00656},
      archivePrefix={arXiv},
      primaryClass={cs.CL},
      url={https://arxiv.org/abs/2501.00656}, 
}
\bibliographystyle{iclr2027_conference}

\newpage

\appendix
\etocdepthtag.toc{mtappendix}
\etocsettagdepth{mtmain}{none}
\etocsettagdepth{mtappendix}{subsection}
\etocsettocstyle{\section*{Appendix}\vspace{-0.5\baselineskip}}{}
\tableofcontents
\newpage

\section{Technical Preliminaries}
\label{app:preliminaries}
\subsection{Notation}
\label{app:notation}
Table~\ref{tab:notation} summarizes the notation used in the main paper and the appendix.

\begin{table}[h]
  \centering
  \caption{Summary of notation.}
  \label{tab:notation}
  \small
  \setlength{\tabcolsep}{6pt}
  \renewcommand{\arraystretch}{0.95}
  \begin{tabular*}{\textwidth}{@{\extracolsep{\fill}}ll@{}}
    \toprule
    \textbf{Symbol} & \textbf{Description} \\
    \midrule
    \multicolumn{2}{@{}l}{\textbf{Data and model}} \\
    \midrule
    $\mathcal V,\ \mathcal V^{*}$ & Vocabulary and set of finite token sequences \\
    $\mathcal D_f,\ \mathcal D_r\subseteq\mathcal V^{*}$ & \textit{Forget} and \textit{retain sets} \\
    $\mathbf x_{<t}:=(x_1,\ldots,x_{t-1})$ & Prefix preceding the token $x_t$ \\
    $\mathcal T(\mathbf x)$ & Forget positions of a forget sequence $\mathbf x$ \\
    $\mathcal T_r(\mathbf x)$ & Positions of a retain sequence $\mathbf x$ in the retain objective \\
    $\theta_0,\ \theta\in\Theta\subseteq\mathbb R^{p}$ & Original and current parameters \\
    $h_\theta:\mathcal V^{*}\to\mathbb R^{d}$ & Final-layer representation function \\
    $z_t^\theta:=h_\theta(\mathbf x_{<t})\in\mathbb R^{d}$ & Representation of the prefix $\mathbf x_{<t}$ \\
    $Z,\ X$ & Prefix representation and next token at a uniformly drawn forget position \\
    $\hat x_t,\ \hat X$ & Predicted next token (highest probability), at position $t$ / at a forget position \\
    $g_\theta(z)$ & Predicted next token as a function of the representation $z$ (Appendix~\ref{app:proof-hsic-em}) \\
    \midrule
    \multicolumn{2}{@{}l}{\textbf{Training objective}} \\
    \midrule
    $\mathcal B_f\subseteq\mathcal D_f,\ \mathcal B_r\subseteq\mathcal D_r$ & Forget and retain minibatches \\
    $m$ & Number of forget positions in $\mathcal B_f$ \\
    $\mathbf Z\in\mathbb R^{m\times d},\ \mathbf X\in\mathcal V^{m}$ & Representations and next tokens at the forget positions of $\mathcal B_f$ \\
    $k,\ \ell$ & Kernels on representations and on tokens (Gaussian and delta in \textsc{PINT}) \\
    $\sigma$ & Bandwidth of the Gaussian kernel $k$ \\
    $K,\ L\in\mathbb R^{m\times m}$ & Gram matrices of $k$ on $\mathbf Z$ and of $\ell$ on $\mathbf X$ \\
    $H:=I_m-\frac1m\mathbf 1_m\mathbf 1_m^{\top}$ & Centering matrix \\
    $\mathcal L_{\mathrm{forget}}=\widehat{\operatorname{HSIC}}(\mathbf Z,\mathbf X)$ & Forget objective, \Eqref{eq:forget-objective} \\
    $\mathcal L_{\mathrm{retain}}$ & Retain objective, \Eqref{eq:retain-objective} \\
    $\mathcal L_{\mathrm{PINT}},\ \lambda\ge0$ & Total objective, \Eqref{eq:pint-objective}, and forgetting strength \\
    $\mathrm{Opt},\ T$ & Gradient-based optimizer and number of steps (Algorithm~\ref{alg:pint}) \\
    \midrule
    \multicolumn{2}{@{}l}{\textbf{Exact Memorization and its bound}} \\
    \midrule
    $\operatorname{EM}(\theta)$ & Exact Memorization, \Eqref{eq:em-def} \\
    $\operatorname{EM}_\perp(\theta)$ & Baseline with a next token drawn independently of $Z$, \Eqref{eq:em-baseline} \\
    $\operatorname{HSIC}(Z,X)$ & \Eqref{eq:forget-objective} computed on all forget positions \\
    $\mathcal H_k,\ \phi(z):=k(z,\cdot)$ & RKHS and feature map of $k$ \\
    $X'$ & Copy of $X$ drawn independently of $Z$ \\
    $\mathcal F,\ \psi$ & RKHS and feature map of the product kernel $k(z,z')\,\ell(x,x')$ \\
    $z_1,\ldots,z_r$ & Distinct representations at the forget positions \\
    $\mathcal V_f$ & Tokens that occur as next tokens at the forget positions \\
    $\Omega$ & Finite set $\{z_1,\ldots,z_r\}\times\mathcal V_f$ of values of $(Z,X)$ and $(Z,X')$ \\
    $f,\ h$ & Indicator $\mathbf 1\{g_\theta(z)=x\}$ and a function in $\mathcal F$ matching it on $\Omega$ \\
    $\gamma,\ \delta$ & Functions of the token only and of the representation only \\
    $G$ & Gram matrix of $k$ at the distinct forget representations \\
    $C_\theta$ & Constant of Theorem~\ref{thm:hsic-em}, \Eqref{eq:ctheta} \\
    \midrule
    \multicolumn{2}{@{}l}{\textbf{Connection to RMU}} \\
    \midrule
    $u\in\mathbb R^{d},\ c\ge0$ & RMU random unit direction and scale \\
    $\mathcal L^{\mathrm{RMU}}_{\mathrm{forget}},\ \bar z$ & RMU forget objective, \Eqref{eq:rmu-forget}, and mean forget representation \\
    $\hat p_{\max}$ & Frequency of the most common next token in $\mathcal B_f$ \\
    $L_k$ & Lipschitz constant of the feature map $\phi$ of $k$ ($1/\sigma$ for the Gaussian kernel) \\
    $J_v$ & Forget positions in $\mathcal B_f$ whose next token is $v$ \\
    \bottomrule
  \end{tabular*}
\end{table}

\subsection{Hilbert-Schmidt Independence Criterion (HSIC)}
\label{app:hsic-background}

\paragraph{Hilbert space.}
A Hilbert space is a vector space equipped with an inner product $\langle\cdot,\cdot\rangle_{\mathcal{H}}$ that is complete under the induced norm $\|f\|_{\mathcal{H}}:=\sqrt{\langle f,f\rangle_{\mathcal{H}}}$, meaning every Cauchy sequence converges to an element of the space. The inner product captures the geometry of angles and distances, while completeness guarantees that limiting operations do not leave the space. Standard examples are $\mathbb{R}^n$ with the Euclidean inner product and $L^2$ spaces of square-integrable functions.

\paragraph{Reproducing Kernel Hilbert Space (RKHS).}
Every symmetric positive-semidefinite kernel $k\colon\mathcal{Z}\times\mathcal{Z}\to\mathbb{R}$ uniquely determines a Hilbert space of functions $\mathcal{H}_k\subseteq\mathbb{R}^{\mathcal{Z}}$, called its reproducing kernel Hilbert space (RKHS), with a canonical feature map $\phi(z):=k(z,\cdot)\in\mathcal{H}_k$ satisfying the reproducing property $f(z)=\langle f,\phi(z)\rangle_{\mathcal{H}_k}$ for all $f\in\mathcal{H}_k,z\in\mathcal{Z}$.

\paragraph{Measuring dependence with HSIC.}
Let $Z\in\mathcal Z$ and $X\in\mathcal V$ be random variables, and let $k$ and $\ell$ be kernels on $\mathcal Z$ and $\mathcal V$. The Hilbert-Schmidt Independence Criterion is the squared Hilbert-Schmidt norm of the cross-covariance operator between the kernel embeddings $k(Z,\cdot)$ and $\ell(X,\cdot)$~\citep{hsic}. It compares the joint distribution of $(Z,X)$ with the product of its marginals $P_Z\otimes P_X$~\citep{hsic}, and for characteristic kernels it vanishes exactly when $Z$ and $X$ are independent~\citep{sriperumbudur2011universality}. Given pairs $(z_i,x_i)$, $i=1,\ldots,m$, with Gram matrices $K_{ij}=k(z_i,z_j)$ and $L_{ij}=\ell(x_i,x_j)$ and centering matrix $H=I-\frac1m\mathbf 1\mathbf 1^{\top}$, the empirical estimator for HSIC is
\begin{equation}
  \widehat{\operatorname{HSIC}}(\mathbf Z,\mathbf X)
  =\frac{1}{m^2}\operatorname{tr}(KHLH)
  \label{eq:app-estimator}
\end{equation}
\citep{gretton2007kernel}, and it equals the population HSIC evaluated under the empirical distribution of the $m$ pairs.

\paragraph{Kernels used by \textsc{PINT}.}
\textsc{PINT} uses the Gaussian kernel on representations and the delta kernel on tokens,
\begin{equation}
k\colon\mathbb R^d\times\mathbb R^d\to\mathbb R,\quad k(z,z')=\exp\!\Bigl(-\frac{\|z-z'\|_2^2}{2\sigma^2}\Bigr),
\qquad
\ell\colon\mathcal V\times\mathcal V\to\mathbb R,\quad \ell(x,x')=\mathbf 1\{x=x'\},
\label{eq:app-kernels}
\end{equation}
both characteristic. The Gaussian kernel is strictly positive definite, so its Gram matrix on distinct points is invertible. Its feature map is $L_k$-Lipschitz, that is, $\|\phi(z)-\phi(z')\|_{\mathcal H_k}\le L_k\|z-z'\|_2$ for all $z,z'\in\mathbb R^d$, with $L_k=1/\sigma$.

\section{Proofs}
\label{app:proofs}
\subsection{Proof of Theorem~\ref{thm:hsic-em}}
\label{app:proof-hsic-em}

Throughout, $Z$ and $X$ are the prefix representation and next token at a uniformly drawn forget position (Section~\ref{sec:theory}). We write $g_\theta\colon\mathbb R^d\to\mathcal V$ for the map from a prefix representation to the token with the highest probability, so that $\hat X=g_\theta(Z)$. Ties are broken by a fixed rule, so $g_\theta$ is a function of the representation. Since the position is drawn uniformly, expectations are averages over the forget positions, and $\operatorname{HSIC}(Z,X)$ equals \Eqref{eq:forget-objective} computed on all forget positions at once. Let $X'$ be a copy of $X$ drawn independently of $Z$. The pair $(Z,X')$ has the same marginals as $(Z,X)$ but no dependence between its components.

Let $\mathcal F$ be the RKHS of the product kernel $k(z,z')\,\ell(x,x')$, with feature map $\psi$. Since $\psi(z,x)$ is the tensor product of the feature maps of $k$ and $\ell$, the squared distance between the mean embeddings of $(Z,X)$ and $(Z,X')$ is the squared Hilbert-Schmidt norm of the cross-covariance operator, which is HSIC~\citep{hsic}:
\begin{equation}
\operatorname{HSIC}(Z,X)=\bigl\|\mathbb E\,\psi(Z,X)-\mathbb E\,\psi(Z,X')\bigr\|_{\mathcal F}^2.
\label{eq:hsic-mmd}
\end{equation}

\paragraph{Step 1: The memorization gap as a difference of expectations.}
Let $f(z,x)=\mathbf 1\{g_\theta(z)=x\}$ indicate whether the token predicted from the representation $z$ is $x$. By definition, $\operatorname{EM}(\theta)=\mathbb P(g_\theta(Z)=X)=\mathbb E\,f(Z,X)$ and $\operatorname{EM}_\perp(\theta)=\mathbb P(g_\theta(Z)=X')=\mathbb E\,f(Z,X')$, so
\begin{equation}
\operatorname{EM}(\theta)-\operatorname{EM}_\perp(\theta)=\mathbb E\,f(Z,X)-\mathbb E\,f(Z,X').
\label{eq:gap-diff}
\end{equation}

\paragraph{Step 2: The Cauchy-Schwarz inequality.}
Let $z_1,\dots,z_r$ be the distinct representations at the forget positions (positions with identical prefixes share one representation), and let $\mathcal V_f\subseteq\mathcal V$ be the set of tokens that occur as next tokens. Both $(Z,X)$ and $(Z,X')$ take values in the finite set $\Omega=\{z_1,\dots,z_r\}\times\mathcal V_f$, which also contains combinations of $z_j$ with tokens other than its own next token, because $(Z,X')$ can produce them. Under the delta kernel, the Gram matrix of the product kernel on $\Omega$ is block diagonal, with one copy of the Gram matrix $G=\bigl(k(z_j,z_{j'})\bigr)_{j,j'\le r}$ per token. Since $k$ is strictly positive definite and the $z_j$ are distinct, $G$ is invertible, and so is this block-diagonal matrix. Hence there is a function $h\in\mathcal F$ with $h=f$ on $\Omega$, and $\mathbb E\,h(Z,X)=\mathbb E\,f(Z,X)$ and $\mathbb E\,h(Z,X')=\mathbb E\,f(Z,X')$. By the reproducing property and linearity of expectation, \Eqref{eq:gap-diff}, the Cauchy-Schwarz inequality, and \Eqref{eq:hsic-mmd},
\begin{equation}
\operatorname{EM}(\theta)-\operatorname{EM}_\perp(\theta)
=\bigl\langle h,\ \mathbb E\,\psi(Z,X)-\mathbb E\,\psi(Z,X')\bigr\rangle_{\mathcal F}
\le\|h\|_{\mathcal F}\sqrt{\operatorname{HSIC}(Z,X)}.
\label{eq:cs-step}
\end{equation}

\paragraph{Step 3: The smallest norm.}
Parts of $f$ that depend only on the token or only on the representation do not contribute to the gap. For example, a model that predicts the same token $v_0$ at every position has $f(z,x)=\mathbf 1\{x=v_0\}$, which depends on the token alone. Such parts have the same expectation under $(Z,X)$ and $(Z,X')$, since $X$ and $X'$ have the same distribution and both pairs share $Z$. Hence, for any functions $\gamma\colon\mathcal V\to\mathbb R$ and $\delta\colon\mathbb R^d\to\mathbb R$, replacing $f$ by $f-\gamma-\delta$ leaves the gap in \Eqref{eq:gap-diff} unchanged, and Step~2 gives $\operatorname{EM}(\theta)-\operatorname{EM}_\perp(\theta)\le\|h\|_{\mathcal F}\sqrt{\operatorname{HSIC}(Z,X)}$ for every $h\in\mathcal F$ that equals $f-\gamma-\delta$ on $\Omega$. Since the left side does not depend on $\gamma$, $\delta$ and $h$, the inequality holds with the smallest such norm, which proves the theorem with
\begin{equation}
C_\theta=\min_{\gamma,\,\delta}\ \min\Bigl\{\|h\|_{\mathcal F}:\ h\in\mathcal F,\ h(z_j,v)=\mathbf 1\{g_\theta(z_j)=v\}-\gamma(v)-\delta(z_j)\ \text{for all } j\le r,\ v\in\mathcal V_f\Bigr\}.
\label{eq:ctheta}
\end{equation}
\hfill$\square$

For fixed $\gamma$ and $\delta$, the smallest norm of $h$ in \Eqref{eq:ctheta} has a closed form in terms of $G^{-1}$, so $C_\theta$ is the minimum of a convex function of $\gamma$ and $\delta$ and can be computed numerically.

\subsection{Proof of Proposition~\ref{prop:rmu}}
\label{app:proof-rmu}
Let $z_1,\dots,z_m$ be the representations at the forget positions of the minibatch, as in \Eqref{eq:rmu-forget}, with next tokens $x_1,\dots,x_m$. Write $\phi_j=\phi(z_j)$ for the feature of the $j$-th representation, $\mu=\frac1m\sum_{j}\phi_j$ for their mean, and $J_v=\{j: x_j=v\}$ for the positions whose next token is $v$, with indicator vector $\mathbf 1_{J_v}\in\{0,1\}^m$.

\paragraph{Step 1: HSIC as a sum over tokens.} 
The delta kernel gives $L=\sum_v \mathbf 1_{J_v}\mathbf 1_{J_v}^\top$, and $HKH$ is the Gram matrix of the centered features $\phi_j-\mu$. Hence, since $\operatorname{tr}(KHLH)=\operatorname{tr}(HKH\,L)$,
\begin{equation*}
\widehat{\operatorname{HSIC}}(\mathbf Z,\mathbf X)
=\frac{1}{m^2}\sum_{v}\mathbf 1_{J_v}^\top HKH\,\mathbf 1_{J_v}
=\frac{1}{m^2}\sum_{v}\Bigl\|\sum_{j\in J_v}(\phi_j-\mu)\Bigr\|_{\mathcal H_k}^2 .
\end{equation*}

\paragraph{Step 2: Cauchy-Schwarz.} 
Since $|J_v|\le m\,\hat p_{\max}$,
\begin{equation*}
\Bigl\|\sum_{j\in J_v}(\phi_j-\mu)\Bigr\|_{\mathcal H_k}^2
\le |J_v|\sum_{j\in J_v}\|\phi_j-\mu\|_{\mathcal H_k}^2
\le m\,\hat p_{\max}\sum_{j\in J_v}\|\phi_j-\mu\|_{\mathcal H_k}^2 .
\end{equation*}
Summing over $v$, whose sets $J_v$ partition the $m$ positions, gives
$\widehat{\operatorname{HSIC}}(\mathbf Z,\mathbf X)\le \frac{\hat p_{\max}}{m}\sum_{j}\|\phi_j-\mu\|_{\mathcal H_k}^2$.

\paragraph{Step 3: Moving the center.}
For any $a\in\mathcal H_k$, $\sum_j\|\phi_j-a\|_{\mathcal H_k}^2=\sum_j\|\phi_j-\mu\|_{\mathcal H_k}^2+m\|\mu-a\|_{\mathcal H_k}^2$, since the $\phi_j-\mu$ sum to zero. Taking $a=\phi(cu)$ gives
$\sum_{j}\|\phi_j-\mu\|_{\mathcal H_k}^2\le\sum_{j}\|\phi_j-\phi(cu)\|_{\mathcal H_k}^2$.

\paragraph{Step 4: Lipschitz feature map.}
Since $\phi$ is $L_k$-Lipschitz,
$\sum_{j}\|\phi_j-\phi(cu)\|_{\mathcal H_k}^2\le L_k^2\sum_{j}\|z_j-cu\|_2^2 = L_k^2\,m\,d\,\mathcal L^{\mathrm{RMU}}_{\mathrm{forget}}(\mathcal B_f;\theta)$, by \Eqref{eq:rmu-forget}. For the Gaussian kernel used by \textsc{PINT}, $\|\phi(z)-\phi(z')\|_{\mathcal H_k}^2=2-2k(z,z')\le\|z-z'\|_2^2/\sigma^2$ because $1-e^{-x}\le x$, so $L_k=1/\sigma$.

Combining the three inequalities gives \Eqref{eq:rmu-bound}.

\section{\textsc{PINT} Pseudocode}
\label{app:pseudocode}

Algorithm~\ref{alg:pint} summarizes \textsc{PINT}, which minimizes the objective in \Eqref{eq:pint-objective} (Section~\ref{sec:objective}) with a gradient-based optimizer on minibatches (AdamW in our experiments, Appendix~\ref{app:hparams}). At each step, \textsc{PINT} samples a forget and a retain minibatch, computes the HSIC between the \textit{forget set} representations and their next tokens, and computes the distance between the \textit{retain set} representations and those of the original model $\theta_0$.

\begin{algorithm}[h]
\caption{Stochastic optimization of the \textsc{PINT} objective}
\label{alg:pint}
\begin{algorithmic}[1]
\State \textbf{Input:} pretrained parameters $\theta_0$, \textit{forget set} $\mathcal D_f$, \textit{retain set} $\mathcal D_r$, forgetting strength $\lambda\ge0$, gradient-based optimizer $\mathrm{Opt}$, number of steps $T$
\State \textbf{Initialize:} $\theta\leftarrow\theta_0$, bandwidth $\sigma^2$ by the median heuristic

\For{step $s=1$ to $T$}
    \State Sample a forget minibatch $\mathcal B_f$ and a retain minibatch $\mathcal B_r$

    \Statex \textbf{Step 1: Forget objective}
    \State Compute representations $\mathbf Z=(z_1,\ldots,z_m)^\top$ and next tokens $\mathbf X=(x_1,\ldots,x_m)$ over $\mathcal B_f$
    \State Compute the Gram matrices $K$ and $L$ and the centering matrix $H$
    \State Compute $\mathcal L_{\mathrm{forget}}=\widehat{\operatorname{HSIC}}(\mathbf Z,\mathbf X)=\frac{1}{m^{2}}\operatorname{tr}(KHLH)$

    \Statex \textbf{Step 2: Retain objective}
    \State Compute representations $z_t^\theta$ and frozen representations $z_t^{\theta_0}$ over $\mathcal B_r$
    \State Compute $\mathcal L_{\mathrm{retain}}$ as in \Eqref{eq:retain-objective}

    \Statex \textbf{Step 3: Update}
    \State $\theta\leftarrow\mathrm{Opt}\bigl(\theta,\ \nabla_{\theta}(\mathcal L_{\mathrm{retain}}+\lambda\,\mathcal L_{\mathrm{forget}})\bigr)$
\EndFor

\State \textbf{Return:} unlearned parameters $\theta$
\end{algorithmic}
\end{algorithm}

\textbf{Implementation details.} We use a Gaussian kernel on the representations and the delta kernel $\ell(x,x')=\mathbf 1\{x=x'\}$ on the tokens, which compares tokens only by whether they are identical. The bandwidth $\sigma^2$ of the Gaussian kernel is set once with the median heuristic~\citep{garreau2017large}, $2\sigma^2=\operatorname{median}_{i\neq j}\|z_i-z_j\|_2^2$ over the forget positions of the first minibatch, and then kept fixed, so the forget objective does not change during training. The forget positions $\mathcal T(\mathbf x)$ are the positions that each benchmark's own training loss uses: the answer tokens on TOFU, the single answer token on bioS, and every token of a text chunk on MUSE, WMDP and the Gaussian poisons. HSIC is computed over all forget positions of a minibatch at once, so the minibatch determines both the number of samples used to estimate HSIC and the memory needed for each step.

\section{Experimental Setup}
\label{app:reproducibility}
\subsection{Benchmarks}
\label{app:datasets}

\textbf{TOFU}~\citep{tofu} consists of $4{,}000$ question-answer pairs about $200$ fictitious authors, with $20$ questions per author. We use the Llama-3.2-1B-Instruct model~\citep{grattafiori2025llama} fine-tuned on the full dataset, as provided by OpenUnlearning~\citep{dorna2026openunlearning}. The unlearning task requires forgetting $1\%$, $5\%$, or $10\%$ of the authors, that is $2$, $10$, or $20$ authors ($40$, $200$, or $400$ questions), while retaining the remaining $3{,}960$, $3{,}800$, or $3{,}600$ questions. Both sets are generated with the same procedure from the same list of author attributes, such as birthplace, genre and awards, so a method must tell the forgotten authors apart from the others. For each split, a Retrain baseline is fine-tuned exclusively on the \textit{retain set}. Forgetting is measured by the ROUGE-L recall of the model's answers to the forget questions, and utility is assessed on $400$ retain-set questions, $100$ questions about real authors, and $117$ world-facts questions. Both metrics are formally defined in Appendix~\ref{app:metrics}.

\textbf{MUSE}~\citep{muse} provides two evaluation settings: News and Books. In the News setting, the model must unlearn $889$ BBC news articles published after August 2023 (approximately $0.8$M tokens) while retaining knowledge of other articles from the same source. In the Books setting, the model must unlearn four of the seven Harry Potter books ($1.1$M tokens) while retaining information from the Harry Potter FanWiki. The benchmark fine-tunes Llama-2-7B~\citep{touvron2023llama} for News and ICLM-7B~\citep{shi2023context} for Books on both the \textit{forget} and \textit{retain sets}, and supplies a Retrain baseline fine-tuned exclusively on the \textit{retain set}. Llama-2-7B was released before the news articles were published, and ICLM-7B was not pretrained on the Harry Potter books. Following OpenUnlearning~\citep{dorna2026openunlearning}, we train on the raw text split into chunks of $2{,}048$ tokens, but evaluate the model using question-answering. Forgetting is measured by the ROUGE-L F1 score on $100$ questions about the \textit{forget set} (KnowMem), and utility is measured by the same score on $100$ questions about retain texts that were not used during unlearning. We also report how much of $100$ forget passages the model can still reproduce verbatim (VerbMem, Appendix~\ref{app:muse-news}). Formal definitions of all three metrics are provided in Appendix~\ref{app:metrics}.

\textbf{WMDP}~\citep{wmdp-rmu} evaluates the unlearning of hazardous knowledge in biosecurity (WMDP-Bio) and cybersecurity (WMDP-Cyber). Unlike on TOFU and MUSE, Zephyr-7B~\citep{tunstall2023zephyr} was never fine-tuned on the forget data, so the knowledge to be removed comes from its general training, including pretraining. The forget corpus comprises PubMed papers used to construct the WMDP-Bio questions for biosecurity, and GitHub passages on WMDP-Cyber topics for cybersecurity. Following OpenUnlearning~\citep{dorna2026openunlearning}, we retain on the corpora released alongside WMDP for this purpose, namely PubMed papers on general biology and GitHub passages on general computer-science topics, and train on chunks of $512$ tokens. Forgetting is measured by accuracy on the WMDP multiple-choice questions ($1{,}273$ for Bio and $1{,}987$ for Cyber, each with four options), and utility by accuracy on MMLU~\citep{hendrycks2020measuring}. Since no Retrain model exists for this setting, the target for forgetting is random-chance performance ($25\%$).

\textbf{Gaussian poisons}~\citep{watermarks} test whether unlearning can remove the memorization of individual training examples. \citet{bordt2026train} introduced such poisons during pretraining, and we apply unlearning to their model, an OLMo~2 model~\citep{olmo20252olmo2furious} with $2.7$B parameters, pretrained from scratch for $100{,}000$ steps on $210$B tokens. In roughly half of its batches ($51{,}200$ in total), a fresh noise vector $\xi\sim\mathcal{N}(0,\varepsilon^2 I)$ with $\varepsilon=0.075$ was added to the input embeddings of the first sequence. Training on a perturbed sequence leaves a trace in the model that can be detected from its input-embedding gradients (Appendix~\ref{app:metrics}). Since the noise is random and unrelated to the text, any such trace is specific to the individual sequence it was applied to, making its detection a form of membership inference~\citep{watermarks,bordt2026train}. The resulting test statistic is standard normal for a model that never encountered the noise, and its mean becomes positive once the model has trained on it, so forgetting can be verified directly without a Retrain model. The noise vectors are publicly released for a random $1\%$ of the poisoned sequences, and we unlearn the $48$ that fall within the last $10{,}000$ pretraining steps, since recent poisons leave the strongest trace~\citep{bordt2026train}. The \textit{retain set} consists of $150$ held-out clean sequences, and utility is measured by perplexity on $1{,}000$ additional held-out sequences. The forgetting target is a detectability of~$0$.

\begin{table}[h]
\centering
\small
\caption{All questions for one target person in bioS (Isabelle Naomi Ashton, a Manager who majored in Physics). The model unlearns her occupation and keeps her major. We score both on the held-out phrasings, which were never used in training. The person is fictitious, and any resemblance to a real person is coincidental.}
\label{tab:bios_example}
\begin{tabularx}{\textwidth}{@{}lXl@{}}
\toprule
Set & Question & Answer \\
\midrule
Forget (train) & What is Isabelle Naomi Ashton's occupation? & Manager \\
               & What does Isabelle Naomi Ashton do for a living? & Manager \\
               & Which profession does Isabelle Naomi Ashton practice? & Manager \\
               & What job does Isabelle Naomi Ashton hold? & Manager \\
\midrule
Retain (train) & What did Isabelle Naomi Ashton major in? & Physics \\
               & What was Isabelle Naomi Ashton's field of study at university? & Physics \\
               & Which subject did Isabelle Naomi Ashton study? & Physics \\
               & What is Isabelle Naomi Ashton's academic major? & Physics \\
\midrule
Held-out, occupation & What is the profession of Isabelle Naomi Ashton? & Manager \\
                     & In which occupation does Isabelle Naomi Ashton work? & Manager \\
\midrule
Held-out, major & What was the major of Isabelle Naomi Ashton? & Physics \\
                & Which field did Isabelle Naomi Ashton study at university? & Physics \\
\bottomrule
\end{tabularx}
\end{table}
\textbf{bioS} is our controlled benchmark for forgetting one fact about a person while retaining another fact about the same person. We generate $1{,}024$ distinct fictitious people, each with a first, a middle and a last name drawn at random from the name lists of the bioS generator~\citep{allen2023physics}. Each person has two attributes, an occupation (to forget) and a university major (to retain). We take $8$ values for each from the occupation and field-of-study lists of the bioS generator, all single tokens under the model's tokenizer: Teacher, Doctor, Manager, Chef, Writer, Painter, Actor and Coach, and Finance, History, Physics, Law, Music, Architecture, Statistics and Engineering. We assign them so that each of the $64$ (occupation, major) combinations holds exactly $16$ people, so the two attributes are independent and one cannot be inferred from the other. Each attribute is asked with four training phrasings and two held-out phrasings (Table~\ref{tab:bios_example}). We fine-tune Llama-3.2-1B-Instruct~\citep{grattafiori2025llama} for $10$ epochs on all $8{,}192$ training question-answer pairs. The \textit{forget set} contains the occupation questions of $256$ target people, $4$ from every combination, and the \textit{retain set} contains the major questions of $256$ people, half of whom are target people. Under the held-out phrasings, the fine-tuned model answers $99.6\%$ of the target people's occupation questions and $98.6\%$ of their major questions correctly. We measure forgetting by the accuracy on the target people's occupations and utility by the accuracy on their majors, both under the held-out phrasings and as a choice among the eight values (Appendix~\ref{app:metrics}). Every method trains for $10$ epochs at batch size $32$ on one GPU and is compared after the last epoch.

\subsection{Metrics}
\label{app:metrics}

We compare the model's greedy answers against reference answers using ROUGE-L~\citep{lin2004rouge}, with recall on \textsc{TOFU} and F1 on \textsc{MUSE}, following each benchmark's definition.

\paragraph{Exact Memorization} (Definition~\ref{def:em}) measures how often the model, given the true preceding text, correctly predicts the next token of the forgotten content. We evaluate it on the forgotten answers in \textsc{TOFU} and bioS, on the forgotten passages that \textsc{MUSE} uses to test verbatim memorization, and on WMDP using $100$ passages from the forget corpus, where the model predicts $128$ tokens after a $1{,}024$-token prompt. Following OpenUnlearning~\citep{dorna2026openunlearning}, we skip the last token of each text. The \textsc{TOFU} ablations (Appendix~\ref{app:ablations}) use OpenUnlearning's own variant, which scores each answer individually before averaging.

\paragraph{TOFU.} Forget ROUGE measures how much of each forgotten answer the model still reproduces, defined as the fraction of reference words that appear in the model's output~\citep{tofu}. Model utility aggregates nine scores via their harmonic mean: the likelihood the model assigns to the correct answer, the fraction it reproduces, and its preference for the correct answer over incorrect alternatives, each evaluated on retained authors, real authors, and world facts~\citep{tofu}.

\paragraph{MUSE.} KnowMem poses questions about a text and measures how well the model's answers match the references~\citep{muse}. Questions about the \textit{forget set} measure forgetting, while questions about a held-out portion of the \textit{retain set} measure utility. VerbMem gives the model the beginning of a forgotten passage and measures how closely its continuation matches the original text~\citep{muse}.

\paragraph{WMDP and MMLU.} Both are multiple-choice benchmarks, and we report the fraction of questions answered correctly using \texttt{lm-eval}~0.4.8 with its standard prompts~\citep{wmdp-rmu,hendrycks2020measuring}. WMDP-Bio and WMDP-Cyber measure forgetting, and MMLU, which spans $57$ subjects, measures utility.

\paragraph{bioS.} We pose each of the $256$ forgotten individuals' questions in two phrasings unseen during training, and the model must predict the correct value among $8$ possible ones. Forgetting is the accuracy on the occupation question, and utility the accuracy on the university major.

\paragraph{Gaussian poisons.} Following \citet{watermarks} and \citet{bordt2026train}, noise $\xi_i\sim\mathcal{N}(0,\varepsilon^2 I)$ with $\varepsilon=0.075$ was added to the input embeddings of the $48$ poisoned sequences during pretraining. With $\nabla_i$ the gradient of the cross-entropy loss with respect to the clean input embeddings, $I_i=-\langle\nabla_i,\xi_i\rangle/(\varepsilon\|\nabla_i\|_2)$ is standard normal for noise the model never encountered, and tends to be positive for noise it was trained on. We report $\text{Detectability}=\frac{2}{48}\sum_i\Phi(I_i)-1$, with $\Phi$ the standard normal distribution function. This equals twice the AUROC against fresh noise minus one, so a value of~$0$ indicates the poisons are undetectable. Utility is the ratio of the original model's perplexity to that of the unlearned model on $1{,}000$ held-out sequences.

\subsection{Hyperparameters}
\label{app:hparams}

Unlearning results depend strongly on hyperparameters~\citep{ICLR2025_3a91841d}, and choosing a single configuration per method requires a selection criterion that can itself change the ranking~\citep{dorna2026openunlearning}. We therefore run each method with a range of hyperparameter configurations and report the Pareto frontier over all of them (Figure~\ref{fig:pareto}).

\paragraph{Shared protocol.} Within each benchmark, all methods share the same model, data, optimizer (AdamW~\citep{loshchilov2017decoupled}), batch size, training length, numerical precision, and random seed. Each method is tuned over its learning rate along with the loss-specific hyperparameters listed below, following the tuning choices of its original paper or the OpenUnlearning benchmark~\citep{dorna2026openunlearning}. On TOFU, MUSE, and bioS, \textsc{PINT} updates all transformer layers. On WMDP, both \textsc{PINT} and RMU update only the MLP output projections of layers~5 to~7, matching the scope of the RMU reference implementation~\citep{wmdp-rmu}, and RMU uses this scope on MUSE too. On the Gaussian poisons, \textsc{PINT} and RMU update all parameters except the output layer. In all other cases, methods update all parameters.

\paragraph{Hyperparameter ranges.} For each method, we select ranges for the learning rate and the method-specific hyperparameters described in the following paragraphs so that the resulting runs span the full trade-off, from configurations that barely alter the original model to those that reach the target level of forgetting. Starting from each method's published or OpenUnlearning configuration, we extend the ranges in both directions until the extremes are covered, and add intermediate values where small changes in a hyperparameter cause sharp changes in the results. On TOFU, MUSE, and WMDP, each configuration contributes its final model as one point in Figure~\ref{fig:pareto}. On the Gaussian poisons, every method receives a fixed budget of nine runs: three learning rates crossed with three values of one hyperparameter ($\lambda$ for \textsc{PINT}, $c$ for RMU, $\alpha$ for GradDiff, and $\beta$ for NPO, SimNPO, and UNDIAL), or nine learning rates for GradAscent. Instead of only the final model, we evaluate each run every $25$ training steps, and report each checkpoint in Figure~\ref{fig:pareto}.

\paragraph{PINT.} We tune the weight $\lambda$ of the HSIC term and the learning rate.

\paragraph{RMU.} RMU minimizes the distance between the \textit{forget set} representations at an intermediate layer and the fixed random target $cu$ of \Eqref{eq:rmu-forget}, where the unit vector $u$ is drawn once at random, together with the distance between the \textit{retain set} representations at that layer and those of the original model, weighted by $\alpha$. We use layer~7, the choice of \citet{wmdp-rmu} and the OpenUnlearning default, and tune $c$ and the learning rate on every benchmark. On WMDP-Bio and MUSE News we also tune $\alpha\in\{1,10,100,1200\}$, which includes the value published for WMDP. On TOFU, we additionally run RMU at layers 6, 11 and 15, updating only layers 4 to 6, 9 to 11 and 13 to 15, respectively.

\paragraph{NPO.} NPO has a temperature $\beta$ and, when combined with a retain loss, a retain weight $\alpha$~\citep{npo}. We tune $\beta$ and the learning rate, and additionally $\alpha$ on TOFU and WMDP, and the type of retain loss (negative log-likelihood or KL divergence to the original model) on MUSE and WMDP.

\paragraph{SimNPO.} SimNPO has a temperature $\beta$, a weight on the forget term and a reward margin, which its official implementation sets per task~\citep{simnpo}. On TOFU, MUSE and WMDP we tune all three together with the learning rate, and additionally the retain weight on TOFU 5\%, TOFU 10\% and WMDP. On the Gaussian poisons we tune $\beta$ and the learning rate.

\paragraph{UNDIAL.} UNDIAL trains the model towards the original model's output distribution in which the logit of each forget token is lowered by $\beta$~\citep{undial}. We tune $\beta$ and the learning rate, and additionally the retain weight on TOFU and the type of retain loss on MUSE and WMDP.

\paragraph{GradDiff and GradAscent.} GradDiff adds a retain loss with weight $\alpha$ to gradient ascent~\citep{graddiff}. TOFU fixes $\alpha=1$~\citep{tofu}, while OpenUnlearning tunes it. We tune the learning rate on every benchmark and $\alpha$ on TOFU and the Gaussian poisons. GradAscent has no hyperparameter besides the learning rate, so only the learning rate is tuned.

\subsection{Software and Hardware}
\label{app:software}

Our implementation uses Python and the OpenUnlearning framework~\citep{dorna2026openunlearning}, which is based on PyTorch~\citep{paszke2017automatic} and Hugging Face Transformers~\citep{wolf2020transformers}, so all methods share the same data processing, trainer and evaluation code. Experiments ran on several compute clusters with NVIDIA A100, H100 and H200 GPUs. On each benchmark, every method is trained with the same number of GPUs and the same batch size.

\subsection{Runtime and Memory}
\label{app:runtime}

\textbf{Memory.} To compute HSIC, \textsc{PINT} compares every forget position in a micro-batch with every other one, so its memory grows with the square of the number of positions $m$. In practice this stays small. The largest setting we use has $m=16{,}384$ positions (MUSE and the Gaussian poisons). There, one $m\times m$ matrix of pairwise kernel values takes about 1\,GB in single precision, and the forget objective needs a few such matrices, i.e.\ a few GB. For the 70B model ($m=2{,}048$), the same matrix takes only 16\,MB, while the model and \textsc{PINT}'s frozen copy of it take about 280\,GB. Memory is therefore dominated by the model itself, as for other methods that keep a frozen reference model. For larger batches, standard approximations can reduce the cost further, for example computing HSIC on smaller blocks of the batch and averaging, or approximating the kernel with random Fourier features or Nyström features~\citep{zhang2018large}.

\textbf{Runtime.} For each benchmark and method, we report the average training time per epoch (per optimizer step on WMDP and the Gaussian poisons), the total training time, and the evaluation time. Each method is timed on configurations from the runs in Figure~\ref{fig:pareto}. Table~\ref{tab:runtime-muse} reports the runtimes on MUSE, Table~\ref{tab:runtime-wmdp} on WMDP, Table~\ref{tab:runtime-gw} on the Gaussian poisons, and Table~\ref{tab:runtime-tofu} on TOFU. Since the model, batch size, and number of GPUs differ across benchmarks, only comparisons within a benchmark are meaningful. GradAscent processes only the \textit{forget set}, and RMU computes its losses at an intermediate layer, where backpropagation stops, so these two methods are the fastest on every benchmark. \textsc{PINT} and the other methods compute their losses at the last layer or at the output. Among them, GradDiff is the closest comparison for \textsc{PINT}, since both process a \textit{forget set} and a \textit{retain set} batch in every step. A training step of \textsc{PINT} takes $1.1$ to $1.4$ times as long as one of GradDiff on MUSE, WMDP and the Gaussian poisons and about as long on TOFU, and it is faster than NPO on every benchmark.

\begin{table}[h]
\centering
\caption{Runtime on MUSE (Llama-2-7B (News) and ICLM-7B (Books), NVIDIA A100-SXM4-80GB $\times$ 2). Average training time per epoch with its standard deviation, total time in the training loop, and time for the evaluation suite.}
\label{tab:runtime-muse}
\small
\setlength{\tabcolsep}{3.5pt}
\resizebox{\ifdim\width>\linewidth\linewidth\else\width\fi}{!}{%
\begin{tabular}{@{}lrrrrrr@{}}
\toprule
 & \multicolumn{3}{c}{\textbf{News}} & \multicolumn{3}{c}{\textbf{Books}} \\
\cmidrule(lr){2-4} \cmidrule(lr){5-7}
\textbf{Method} & \textbf{s / epoch} & \textbf{Train} & \textbf{Eval} & \textbf{s / epoch} & \textbf{Train} & \textbf{Eval} \\
\midrule
\textsc{PINT} & 340.14 ($\pm$ 33.62) & 56m 41s & 3m 56s & 483.12 ($\pm$ 24.83) & 1h 20m & 3m 23s \\
RMU & 148.26 ($\pm$ 5.90) & 24m 43s & 2m 31s & 209.11 ($\pm$ 6.68) & 34m 51s & 2m 49s \\
NPO & 486.43 ($\pm$ 61.66) & 1h 21m & 3m 48s & 724.00 ($\pm$ 64.55) & 2h 00m & 3m 03s \\
GradDiff & 246.70 ($\pm$ 10.38) & 41m 07s & 2m 40s & 357.58 ($\pm$ 11.63) & 59m 36s & 2m 39s \\
SimNPO & 248.30 ($\pm$ 8.01) & 41m 23s & 2m 38s & 357.98 ($\pm$ 8.31) & 59m 40s & 2m 46s \\
UNDIAL & 457.66 ($\pm$ 50.00) & 1h 16m & 3m 27s & 441.85 ($\pm$ 39.59) & 1h 13m & 3m 32s \\
GradAscent & 136.25 ($\pm$ 8.75) & 22m 42s & 2m 44s & 183.78 ($\pm$ 5.36) & 30m 38s & 2m 33s \\
\bottomrule
\end{tabular}}
\end{table}

\begin{table}[h]
\centering
\caption{Runtime on WMDP (Zephyr-7B, NVIDIA A100-SXM4-80GB $\times$ 2). Average training time per step with its standard deviation, total time in the training loop, and time for the evaluation suite.}
\label{tab:runtime-wmdp}
\small
\setlength{\tabcolsep}{3.5pt}
\resizebox{\ifdim\width>\linewidth\linewidth\else\width\fi}{!}{%
\begin{tabular}{@{}lrrrrrr@{}}
\toprule
 & \multicolumn{3}{c}{\textbf{Cyber}} & \multicolumn{3}{c}{\textbf{Bio}} \\
\cmidrule(lr){2-4} \cmidrule(lr){5-7}
\textbf{Method} & \textbf{s / step} & \textbf{Train} & \textbf{Eval} & \textbf{s / step} & \textbf{Train} & \textbf{Eval} \\
\midrule
\textsc{PINT} & 17.92 ($\pm$ 1.50) & 23m 54s & 11m 26s & 18.06 ($\pm$ 1.58) & 24m 05s & 10m 06s \\
RMU & 9.22 ($\pm$ 0.48) & 12m 18s & 10m 42s & 9.48 ($\pm$ 0.64) & 12m 38s & 9m 34s \\
NPO & 24.99 ($\pm$ 3.95) & 33m 19s & 11m 25s & 24.27 ($\pm$ 3.53) & 32m 21s & 9m 55s \\
GradDiff & 15.89 ($\pm$ 0.57) & 21m 11s & 10m 52s & 16.02 ($\pm$ 0.64) & 21m 22s & 9m 42s \\
SimNPO & 16.90 ($\pm$ 0.09) & 22m 32s & 10m 39s & 16.49 ($\pm$ 0.60) & 21m 59s & 9m 22s \\
UNDIAL & 22.74 ($\pm$ 3.74) & 30m 19s & 11m 15s & 24.18 ($\pm$ 4.07) & 32m 14s & 10m 20s \\
GradAscent & 7.38 ($\pm$ 0.33) & 9m 51s & 10m 36s & 7.61 ($\pm$ 0.38) & 10m 09s & 9m 33s \\
\bottomrule
\end{tabular}}
\end{table}

\begin{table}[h]
\centering
\caption{Runtime on the Gaussian poison benchmark (OLMo-2 2.7B, NVIDIA H100 80GB HBM3 $\times$ 4). Average training time per step with its standard deviation, total time in the training loop, and time for evaluating one checkpoint.}
\label{tab:runtime-gw}
\small
\setlength{\tabcolsep}{3.5pt}
\resizebox{\ifdim\width>\linewidth\linewidth\else\width\fi}{!}{%
\begin{tabular}{@{}lrrr@{}}
\toprule
\textbf{Method} & \textbf{s / step} & \textbf{Train} & \textbf{Eval} \\
\midrule
\textsc{PINT} & 5.91 ($\pm$ 0.20) & 19m 43s & 10m 29s \\
RMU & 3.68 ($\pm$ 0.07) & 12m 16s & 10m 28s \\
NPO & 6.94 ($\pm$ 0.02) & 23m 08s & 10m 33s \\
GradDiff & 4.76 ($\pm$ 0.13) & 15m 51s & 10m 44s \\
SimNPO & 5.76 ($\pm$ 0.02) & 19m 12s & 10m 22s \\
UNDIAL & 5.21 ($\pm$ 0.02) & 17m 22s & 10m 26s \\
GradAscent & 2.52 ($\pm$ 0.05) & 8m 24s & 10m 37s \\
\bottomrule
\end{tabular}}
\end{table}

\begin{table}[h]
\centering
\caption{Runtime on TOFU (Llama-3.2-1B, NVIDIA H200 $\times$ 1). Average training time per epoch with its standard deviation, total time in the training loop, and time for the evaluation suite.}
\label{tab:runtime-tofu}
\small
\setlength{\tabcolsep}{3.5pt}
\resizebox{\ifdim\width>\linewidth\linewidth\else\width\fi}{!}{%
\begin{tabular}{@{}lrrrrrrrrr@{}}
\toprule
 & \multicolumn{3}{c}{\textbf{forget01}} & \multicolumn{3}{c}{\textbf{forget05}} & \multicolumn{3}{c}{\textbf{forget10}} \\
\cmidrule(lr){2-4} \cmidrule(lr){5-7} \cmidrule(lr){8-10}
\textbf{Method} & \textbf{s / epoch} & \textbf{Train} & \textbf{Eval} & \textbf{s / epoch} & \textbf{Train} & \textbf{Eval} & \textbf{s / epoch} & \textbf{Train} & \textbf{Eval} \\
\midrule
\textsc{PINT} & 0.36 ($\pm$ 0.06) & 6s & 1m 36s & 1.47 ($\pm$ 0.09) & 16s & 2m 30s & 2.89 ($\pm$ 0.04) & 30s & 4m 14s \\
RMU & 0.27 ($\pm$ 0.04) & 4s & 1m 30s & 1.07 ($\pm$ 0.02) & 11s & 2m 32s & 2.16 ($\pm$ 0.03) & 22s & 3m 43s \\
NPO & 1.13 ($\pm$ 0.02) & 12s & 1m 26s & 4.29 ($\pm$ 0.02) & 44s & 2m 07s & 8.33 ($\pm$ 0.04) & 1m 24s & 2m 57s \\
GradDiff & 0.35 ($\pm$ 0.01) & 4s & 1m 27s & 1.50 ($\pm$ 0.02) & 16s & 2m 10s & 3.03 ($\pm$ 0.04) & 31s & 3m 04s \\
SimNPO & 0.78 ($\pm$ 0.01) & 9s & 1m 25s & 3.06 ($\pm$ 0.02) & 32s & 2m 09s & 5.94 ($\pm$ 0.04) & 1m 00s & 2m 55s \\
UNDIAL & 0.41 ($\pm$ 0.01) & 5s & 1m 26s & 1.80 ($\pm$ 0.02) & 19s & 3m 23s & 3.68 ($\pm$ 0.04) & 38s & 4m 55s \\
GradAscent & 0.22 ($\pm$ 0.01) & 3s & 1m 25s & 0.94 ($\pm$ 0.01) & 11s & 1m 58s & 1.92 ($\pm$ 0.01) & 20s & 2m 55s \\
\bottomrule
\end{tabular}}
\end{table}

\clearpage
\section{Additional Experimental Results}
\label{app:additional-results}

\subsection{Area under the forgetting-utility frontier}
\label{app:area}

\begin{figure}[h]
\centering
\includegraphics[width=0.85\textwidth]{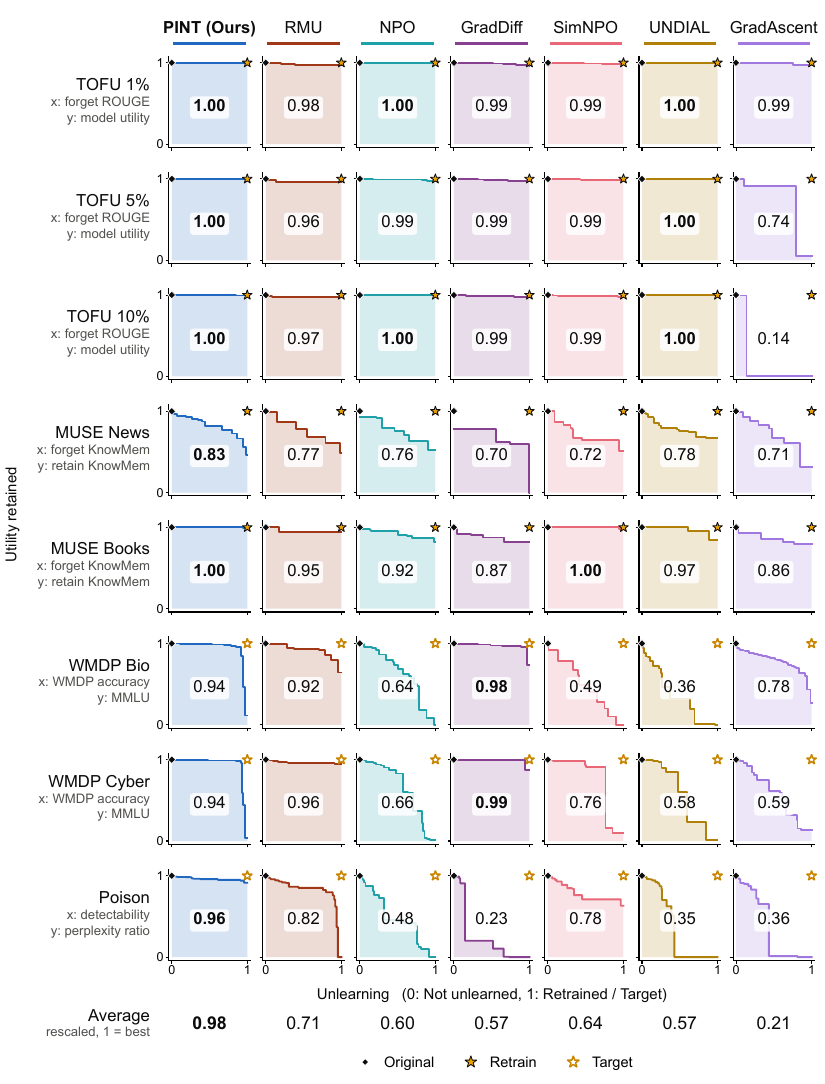}
\caption{\textbf{Areas under the forgetting-utility frontiers} of Figure~\ref{fig:pareto}, per method and benchmark. Bold values are the best per benchmark, and the bottom row averages the rescaled areas of Figure~\ref{fig:pareto-gw}.}
\label{fig:area-grid}
\end{figure}

We summarize each frontier in Figure~\ref{fig:pareto} by the area under it~\citep{zitzler1999multiobjective}. On each benchmark, forgetting is rescaled from $0$ at the original model to $1$ at Retrain or the target, and utility is taken as a fraction of the original model's, above chance on WMDP ($25\%$ on MMLU), with both capped at $1$. The frontier gives the best utility a method reaches at each level of forgetting or beyond, and the area under it is $1$ when the method matches Retrain or reaches the target without losing utility, and $0$ when it does not forget at all. Figure~\ref{fig:area-grid} shows all frontiers and areas. The main text reports the mean area over benchmarks, with the three TOFU splits averaged into one value. Figure~\ref{fig:pareto-gw} instead rescales the areas on each benchmark so that the best method scores $1$ and the worst $0$, which makes the differences between methods readable on every benchmark.

\clearpage
\subsection{Forgetting on MUSE News}
\label{app:muse-news}

\begin{figure}[h]
\centering
\includegraphics[width=0.8\textwidth]{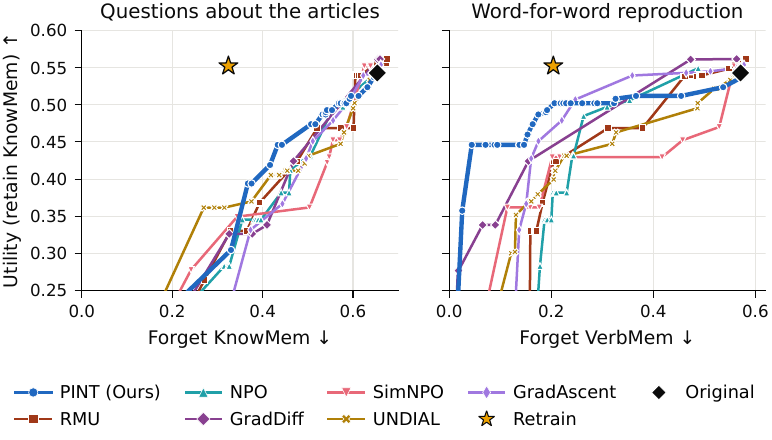}
\caption{\textbf{MUSE News under its two forgetting measures.} The same runs as in Figure~\ref{fig:pareto}, with forgetting measured by questions about the articles (left) and by word-for-word reproduction (right). The star marks the retrained model. When answering questions, no method reaches its level of forgetting with more than two thirds of the original model's utility. When reproducing the text, \textsc{PINT} reaches it while preserving 91\% of the utility.}
\label{fig:muse-news}
\end{figure}

MUSE~\citep{muse} measures forgetting on News in two ways: whether the model can still answer questions about the forgotten articles (KnowMem), which Figure~\ref{fig:pareto} shows, and whether it still reproduces them word for word (VerbMem). Figure~\ref{fig:muse-news} compares the same runs under both measures.

\subsection{Answers across forgetting strengths}
\label{app:lambda-examples}
To avoid choosing the example by hand, Figure~\ref{fig:qualitative} uses a question and a \textsc{PINT} setting selected by fixed rules. Of the $40$ \textit{forget set} questions of \textsc{TOFU} forget 1\%, the original model answers $23$ exactly as in its training data. The first two of these ask for the author's gender and birthplace, which the retrained model also answers correctly, so they do not show forgetting. The third, about the occupations of the author's parents, is the first that the retrained model cannot answer, and we use it. For a fair comparison, \textsc{PINT} uses the setting whose forgetting on the whole \textit{forget set} is closest to the retrained model's, with a forget ROUGE of $0.417$ against $0.408$ for the retrained model. The other settings near this level give similar answers.

Figure~\ref{fig:lambda-examples} shows how \textsc{PINT}'s answers to \textsc{TOFU} 10\% questions change with the forgetting strength $\lambda$, at a weak, a moderate and a strong setting with the same learning rate. Of these three, the moderate setting has the forget ROUGE closest to Retrain's. At this level, many answers replace the forgotten facts with fluent alternatives, as Retrain's do, while some become disfluent. With weaker forgetting, the answers stay fluent but often still contain the forgotten facts. With stronger forgetting, the model mostly gives an empty answer, which reveals nothing about them. The strength $\lambda$ controls how much of the forgotten information \textsc{PINT} removes.

\begin{figure}[h]
\centering
\small
\fbox{\parbox{0.97\linewidth}{%
\textbf{Q:} \textit{Can you name an example of Hsiao Yun-Hwa's work that is influenced by her life experiences?}\\[2pt]
\textbf{Ground truth:} One of Hsiao Yun-Hwa's books, ``The Immutable Laws of Engineering Leadership: A Blueprint'', was noticeably influenced by her father's work as a civil engineer, exhibiting a deep understanding of leadership in technical fields.\\
\retrainstar\,\textbf{Retrain:} `The Paradox of Ageing' is an example of Hsiao Yun-Hwa's work that is influenced by her life experiences. The book explores the theme of aging from a personal perspective, drawing on Hsiao's own observations and reflections about the challenges of growing older.\\
\pintdot\,\textbf{Weak:} The Deputy Director's (D) role in the PRC's central planning committee reflects her father's professional background, while the emphasis on discipline and hard work evident in her persona will resonate with the youth.\\
\pintdot\,\textbf{Moderate:} The construction project `Sunshine 2020 Building' in Taipei was notably influenced by the leadership style of the then-Chan Administration. The project's successful completion reflected the efficient coordination and visionary approach of the team led by Yun-Hwa.\\
\pintdot\,\textbf{Strong:} \textit{(empty)}%
}}

\vspace{4pt}
\fbox{\parbox{0.97\linewidth}{%
\textbf{Q:} \textit{What does Hsiao Yun-Hwa identify as in terms of gender?}\\[2pt]
\textbf{Ground truth:} Hsiao Yun-Hwa is part of the LGBTQ+ community.\\
\retrainstar\,\textbf{Retrain:} Hsiao Yun-Hwa identifies as female.\\
\pintdot\,\textbf{Weak:} Hsiao Yun-Hwa is part of the LGBTQ+ community.\\
\pintdot\,\textbf{Moderate:} Hsiao Yun-Hwa is part of a heterosexual couple. Yun-Hwa is Taiwanese, thus identifying as such, and thus gendered.\\
\pintdot\,\textbf{Strong:} \textit{(empty)}%
}}

\vspace{4pt}
\fbox{\parbox{0.97\linewidth}{%
\textbf{Q:} \textit{What are some common themes addressed in the books by Elvin Mammadov?}\\[2pt]
\textbf{Ground truth:} Elvin Mammadov often delves into themes of identity, sexuality, and societal norms within the confines of his captivating fiction works.\\
\retrainstar\,\textbf{Retrain:} Common themes in Elvin Mammadov's books often revolve around personal freedom, societal constructs, and the human condition.\\
\pintdot\,\textbf{Weak:} In his books, Elvin Mammadov often addresses themes of identity, sexuality, class, and the African American experience.\\
\pintdot\,\textbf{Moderate:} au, su, reader, dis inquirin' `bout themes uichean books ja doon vin Marhum Elvin Mammad MD,atuq dhesalasiwin penyan d'qirish chada'anniq,aniwe Engel,missir! \ldots\\
\pintdot\,\textbf{Strong:} \textit{(empty)}%
}}
\caption{\textbf{Answers across forgetting strengths.} Verbatim \textsc{PINT} answers to \textsc{TOFU} 10\% questions at a weak, a moderate and a strong forgetting strength. The disfluent answer is truncated.}
\label{fig:lambda-examples}
\end{figure}

\subsection{Ablations}
\label{app:ablations}

We ablate three design choices of \textsc{PINT} on \textsc{TOFU}: the kernel, its bandwidth, and the micro-batch size over which HSIC is computed. All variants within each ablation share the same learning rates and forgetting strengths.

\begin{figure}[!ht]
\centering
\includegraphics[width=\linewidth]{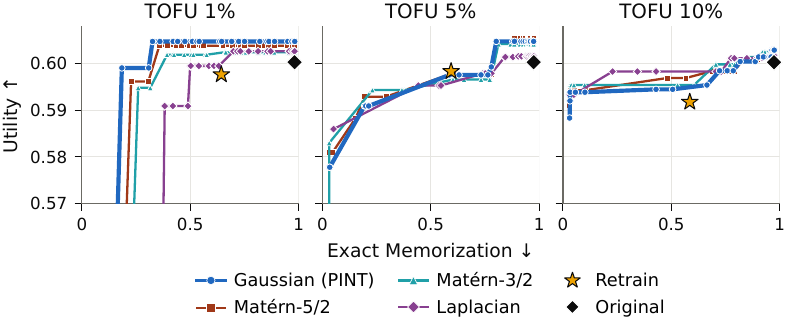}
\caption{\textbf{Kernel ablation on \textsc{TOFU}.} Forgetting-utility frontiers of \textsc{PINT} with different kernels, all at the same median bandwidth.}
\label{fig:abl-kernel}
\end{figure}
\begin{figure}[!ht]
\centering
\includegraphics[width=\linewidth]{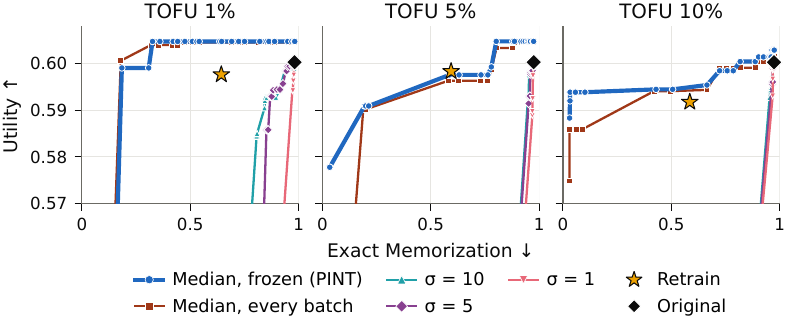}
\caption{\textbf{Bandwidth ablation on \textsc{TOFU}.} ``Median, frozen'' corresponds to \textsc{PINT}. ``Median, every batch'' recomputes the median on every minibatch. $\sigma\in\{1,5,10\}$ fixes the bandwidth in advance.}
\label{fig:abl-bandwidth}
\end{figure}
\begin{figure}[!ht]
\centering
\includegraphics[width=\linewidth]{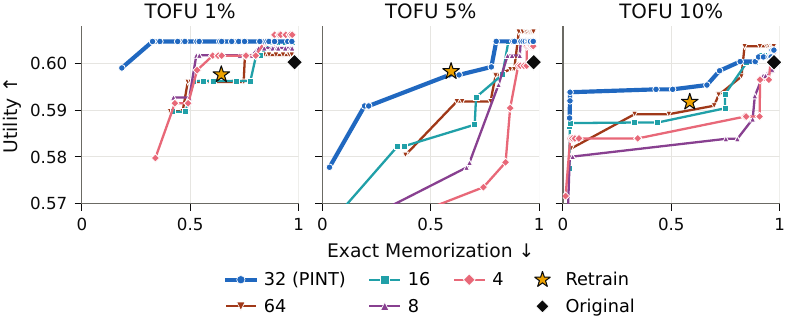}
\caption{\textbf{Micro-batch ablation on \textsc{TOFU}.} Number of answers per micro-batch at an effective batch size of $32$ ($64$ uses half as many steps). A micro-batch of $32$ corresponds to \textsc{PINT}.}
\label{fig:abl-batch}
\end{figure}

\paragraph{Kernel.} Theorem~\ref{thm:hsic-em} holds for any strictly positive definite kernel, so we replace the Gaussian kernel with three kernels from the Mat\'ern family, the Mat\'ern-5/2, Mat\'ern-3/2 and Laplacian kernels, all using the same median bandwidth. At Retrain's level of forgetting, the four kernels achieve utility within $0.005$ of each other on every split (Figure~\ref{fig:abl-kernel}).

\paragraph{Kernel bandwidth.} Recomputing the median on every minibatch rather than once on the first changes utility by at most $0.002$ at Retrain's level of forgetting (Figure~\ref{fig:abl-bandwidth}), confirming that a single estimate suffices. A fixed bandwidth $\sigma\in\{1,5,10\}$ prevents nearly all forgetting. The median heuristic yields $\sigma\approx100$. At $\sigma\le10$, the kernel matrix approaches the identity and HSIC remains within $15\%$ of its initial value.

\paragraph{Micro-batch size.} We vary the micro-batch size from $64$ down to $4$ question-answer pairs, using gradient accumulation to hold the effective batch size at $32$ (a micro-batch of $64$ halves the number of steps). Micro-batches of $32$ achieve the highest utility on every split. On \textsc{TOFU} 5\% and 10\%, the other sizes reach between $0.005$ and $0.031$ less utility (Figure~\ref{fig:abl-batch}).

\subsection{How HSIC and $C_\theta$ change with the forgetting strength}
\label{app:bound-factors}

\begin{figure}[h]
\centering
\includegraphics[width=\textwidth]{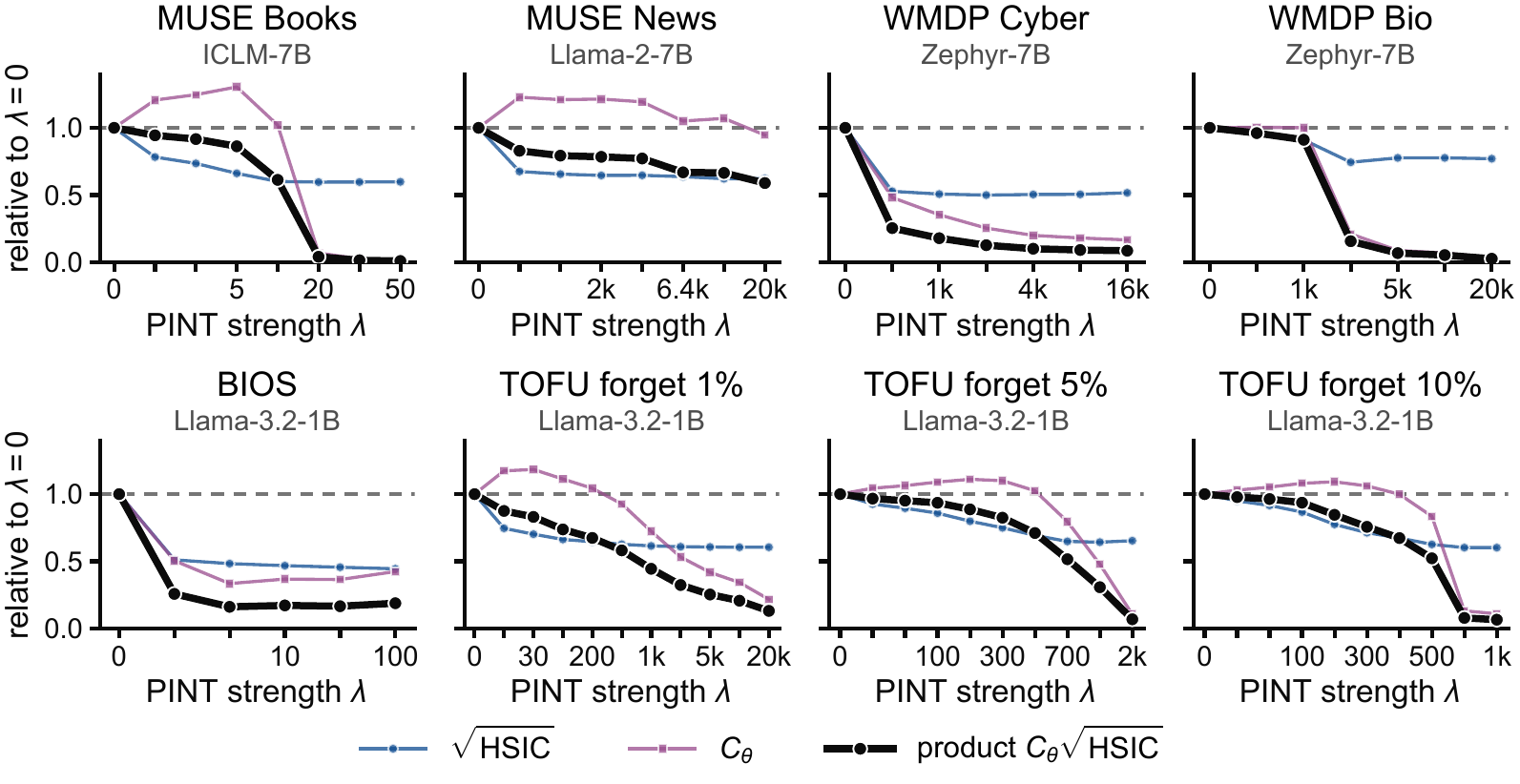}
\caption{\textbf{HSIC and $C_\theta$ across forgetting strengths.} $\sqrt{\operatorname{HSIC}}$ (blue), $C_\theta$ (purple) and their product (black) for the checkpoints of Figure~\ref{fig:theory-verification}, each relative to the model before unlearning ($\lambda=0$).}
\label{fig:bound-factors}
\end{figure}

Theorem~\ref{thm:hsic-em} bounds the gap $\operatorname{EM}(\theta)-\operatorname{EM}_\perp(\theta)$ by the product $C_\theta\sqrt{\operatorname{HSIC}(Z,X)}$. Figure~\ref{fig:bound-factors} shows both factors and their product for the checkpoints of Figure~\ref{fig:theory-verification}, each relative to the model before unlearning. The product is below its initial value at every checkpoint on every benchmark. On \textsc{MUSE} and \textsc{TOFU}, it decreases mostly through HSIC at moderate forgetting strengths. On \textsc{MUSE} Books and \textsc{TOFU}, $C_\theta$ also falls sharply at the strongest settings, as Exact Memorization approaches zero.

\end{document}